\documentclass[10pt,twocolumn,letterpaper]{article}

\usepackage{iccv}
\usepackage{times}
\usepackage{epsfig}
\usepackage{graphicx}
\usepackage{amsmath}
\usepackage{amssymb}
\usepackage{pict2e}
\usepackage{tikz}

\usepackage{multirow}
\usepackage[percent]{overpic}
\usepackage{bm}

\DeclareMathOperator*{\argmax}{arg\,max}

\newcommand{\decparams}{\boldsymbol\theta}
\newcommand{\encparams}{\boldsymbol\phi}

\usepackage[pagebackref=true,breaklinks=true,letterpaper=true,colorlinks,bookmarks=false]{hyperref}

\iccvfinalcopy 

\def\iccvPaperID{12146} 

\ificcvfinal\fi

\newcommand{\figpart}[1]{\mbox{\textbf{(#1)}}\xspace}
\newcommand{\miniparagraph}[1]{\vspace{1.2mm}\noindent\textbf{#1.}}
\newcommand{\miniheadline}[1]{\vspace{1.2mm}\noindent\textit{#1.}}

\def\eg{\emph{e.g}\onedot} 
\def\ie{\emph{i.e}\onedot} 
 
 \def\vs{\emph{vs}\onedot}
\def\wrt{w.r.t\onedot} 
\def\etal{\emph{et~al}\onedot}

\newcommand{\HAE}{\mbox{\textsc{HAE}}\xspace}
\newcommand{\HAEs}{\mbox{\textsc{HAEs}}\xspace}

\newcommand{\HierarchicalAutoencs}{\mbox{\textsc{Hierarchical Autoencoders}}\xspace}

\newcommand{\HVAE}{\mbox{\textsc{HVAE}}\xspace}
\newcommand{\HVAEs}{\mbox{\textsc{HVAEs}}\xspace}

\newcommand{\HierarchicalVAE}{\mbox{\textsc{Hierarchical VAE}}\xspace}
\newcommand{\HierarchicalVAEs}{\mbox{\textsc{Hierarchical VAEs}}\xspace}

\newcommand{\UNet}{\mbox{\textsc{U-Net}}\xspace}
\newcommand{\UNets}{\mbox{\textsc{U-Nets}}\xspace}
\newcommand{\DoubleDIP}{\mbox{Double-DIP}\xspace}

\newcommand{\muSplit}{\mbox{$\mu\text{Split}$}\xspace}
\newcommand{\textmathbf}[1]{\textbf{\boldmath#1}}

\newcommand{\PaviaATN}{\mbox{\textit{PaviaATN}}\xspace}
\newcommand{\HagenEtAl}{\mbox{\textit{Hagen~\etal}}\xspace}
\newcommand{\SinosoidalCritters}{\mbox{\textit{SinosoidalCritters}}\xspace}
\newcommand{\leanLC}{\mbox{\textit{Lean-LC}}\xspace}
\newcommand{\regularLC}{\mbox{\textit{Regular-LC}}\xspace}
\newcommand{\deepLC}{\mbox{\textit{Deep-LC}}\xspace}

\newcommand\figTeaser{
\begin{figure}[t]
\centering
\includegraphics[width=\linewidth]{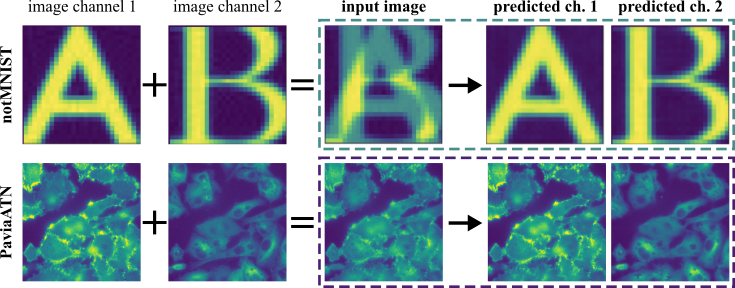}
\caption{Splitting of superimposed image channels.
The input image is the sum of two image channels, each channel containing structures from one given object class.
The task of \muSplit is to identify and split the structures superimposed in the given input image (dashed rectangles).
}
\label{fig:teaser}
\end{figure} 
}

\newcommand\figArchitecture{
\begin{figure*}[t]
\centering
\begin{overpic}[width=.95\textwidth, tics=5, 
                    ]
                    {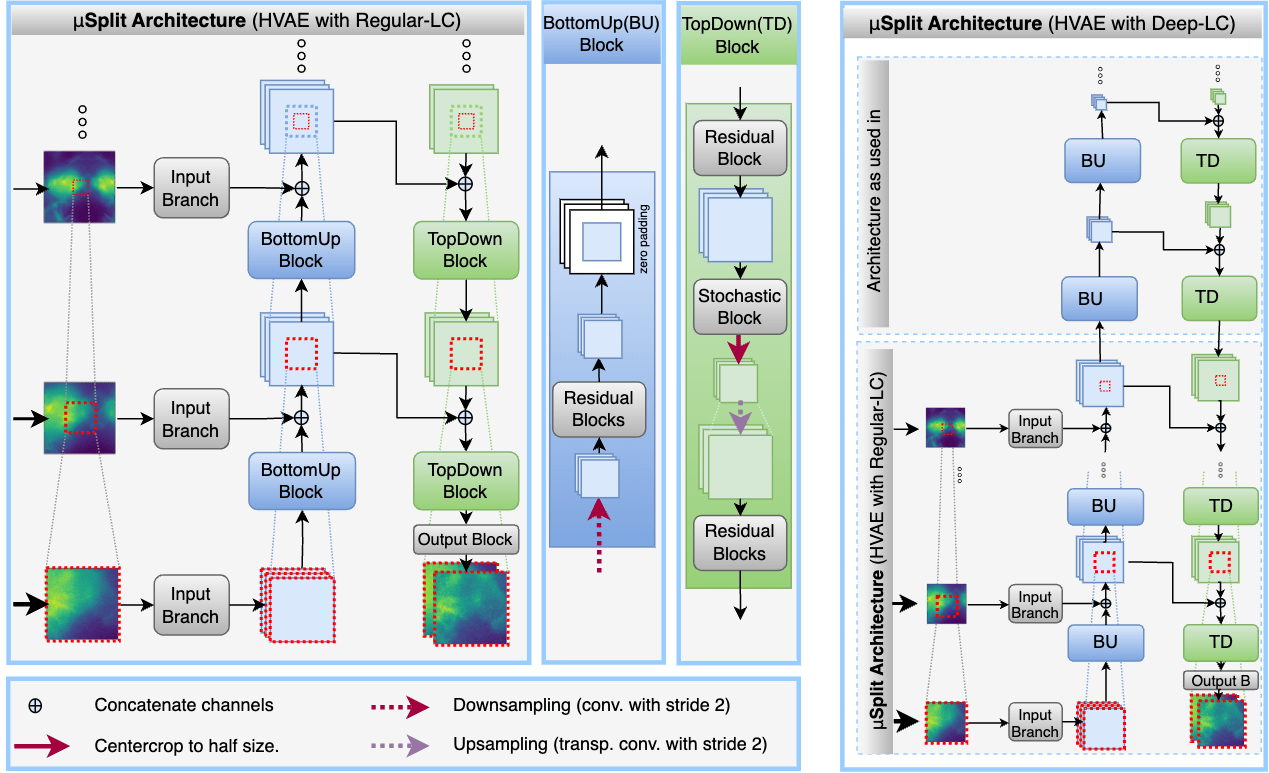}
        \put (-2.7, 61) {\figpart{a}}
        \put (63, 61) {\figpart{b}}
        \put (68.6, 53) {\scriptsize \rotatebox{90}{\cite{Prakash2020-wr}}}
        \put (87, 47.8) {\scriptsize \cite{Prakash2020-wr}}
        \put (86.9, 36.9) {\scriptsize \cite{Prakash2020-wr}}

        \put (96.1, 47.8) {\scriptsize \cite{Prakash2020-wr}}
        \put (96.1, 36.9) {\scriptsize \cite{Prakash2020-wr}}

        \put (5,17.2) {$x_p$}
        \put (5,32) {$x_{(p,1)}$}
        
\end{overpic}
\caption{Network architecture of \muSplit. 
In \figpart{a}, we show the network architecture employed by \regularLC.
The input (left side) consists of a core image patch $x_p$, together with downscaled version of the patch surroundings -- the lateral context (LC).
We show the area corresponding to the original patch as red dotted box throughout the figure.
\figpart{b}~The network architecture of \deepLC. 
The architecture used in~\cite{Prakash2020-wr} is stacked on top of the \regularLC architecture shown in (a). 
Note that this is only possible because the latent space in \regularLC retained the spatial dimensions of all layers by means of using the proposed LC. 
\textit{Note:} a sketch of the \leanLC architecture, our third LC variant, can be found in the Supp. Figure~\textit{S.1}.
}
\label{fig:overall_architecture}
\end{figure*}
}

\newcommand\figPaddingFigures{
\begin{figure}[t]
    \centering
    \begin{overpic}[width=.995\columnwidth, tics=5, 
                    ]
                    {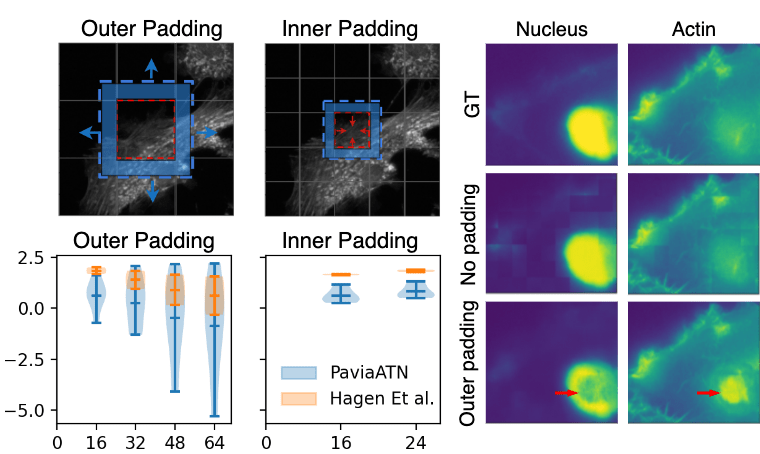}
        \put (3.0, 54) {\figpart{a}}
        \put (3.0, 28) {\figpart{b}}
        \put ( 59, 54) {\figpart{c}}
        
    \end{overpic}
    \caption{\textbf{Strategies for tiled predictions.}
    \figpart{a}~The difference between Inner and Outer Padding. 
    The blue dashed rectangle represents one patch used for tiled predictions.
    For each cell in the faint gray grid superimposed on the input image one such patch exists.
    The red dashed rectangle represents the center-crop region used to tile the final prediction of the entire input image.
    The blue shaded area is therefore the part of the patch that overlaps with neighboring patches, \ie it is the padding area for the red rectangle.
    \textit{Outer Padding} uses a tile size equivalent to the training patch size and introduces overlap by enlarging the patch being fed to the network. 
    \textit{Inner Padding}, in contrast, maintains the original patch size, and uses only an inner crop to tile the given input image. 
    \figpart{b}~Percentange variation (of PSNR measurements) when using different amounts of Outer or Inner padding (for \HAE and \HVAE vanilla setups using a patch size of 64). 
    For varying amounts of padding (x-axis), we plot how $6$ data points for the \PaviaATN data ($3\text{ tasks}*2=6$) and $2$ data points for \HagenEtAl data ($1$ task) are distributed.
    Note how distributions for Inner Padding are consistently better.
    \figpart{c}~Using Outer Padding, predictions are performed on patches larger than the ones used during training, leading to out-of-distribution (OOD) inputs and therefore to inferior predictions (red arrows). First and second row are the ground truth and prediction made without any padding respectively.
    See Supp.~Figure~\textit{S.3} for more examples.
    }
    \label{fig:padding_figures}
\end{figure}
}

\newcommand\figSynthetic{
\begin{figure*}[tb]
\centering
    \begin{overpic}[width=.95\textwidth, tics=5] 
            {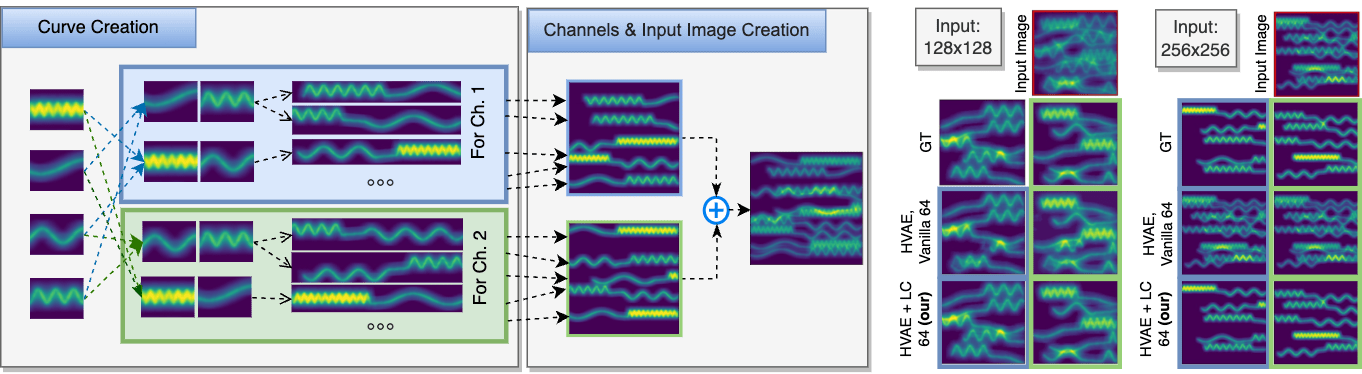}
        \put (-2.5, 26) {\figpart{a}}
        \put (64.5, 26) {\figpart{b}}

        \put (1.8, 21.2) {\tiny Frequencies}
        \put (9.7, 23) {\tiny Freq. pairs, \ie, Critters}
        \put (21.8, 23) {\tiny Connecting these Freq. pairs}
        \put (43.3, 21.8) {\tiny Ch1. Image}
        \put (43.3, 11.5) {\tiny Ch2. Image}
        \put (56.9, 16.5) {\tiny Input Image}
        \put (72, 9) { \color{red}\circle{2}}

    \end{overpic}
\caption{
The synthetic \SinosoidalCritters dataset is designed in such a way that large lateral image context is needed in order to perform correct channel splitting. 
\figpart{a}~A schema illustrating how we created the \SinosoidalCritters dataset. A detailed description is provided in Section~\ref{sec:data}.
\figpart{b}~We show two sample \SinosoidalCritters input images (row 1) of size $128\times 128$ and $256\times 256$ pixels and the two channels that created them (row 2), respectively.
Below, we show the decomposition results obtained with a trained vanilla \HVAE with input patch size 64 (row 3), and results obtained with the same architecture but using \regularLC (row 4).
To recognise which critter is depicted and assign it to a channel, the network has to see both wave forms, hence requiring long range lateral image context. 
}
\label{fig:datagen}
\label{fig:critters}
\end{figure*}
}

\newcommand\figPlots{
\begin{figure*}[tb]
    \centering
    \begin{overpic}[width=.38\textwidth]{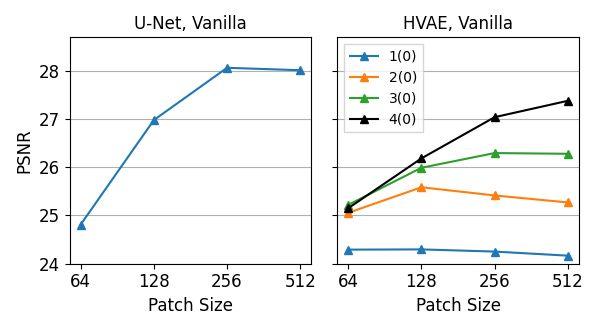}
        \put (0, 50) {\figpart{a}}
    \end{overpic}
    \hspace{5mm}
    \begin{overpic}[width=.58\textwidth]{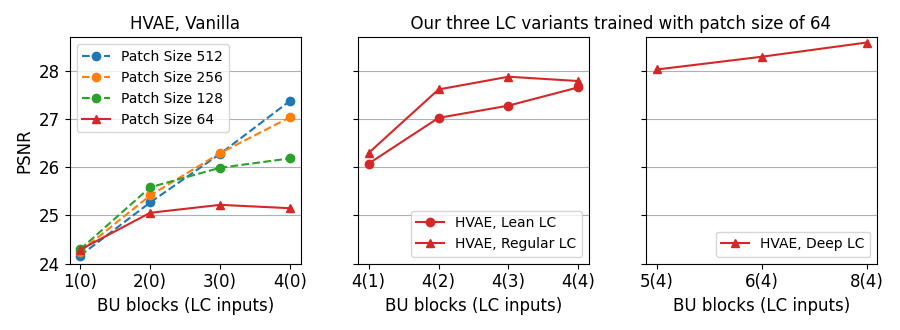}
        \put (0, 32) {\figpart{b}}
        \put(34.3,13.5){\tikz \draw[-to,dashed,red](0,0)--(0.5,0.4);}
        \put(66,27.2){\tikz \draw[-to,dashed,red](0,0)--(0.5,0.12);}

    \end{overpic}
    \caption{\textmathbf{Benefits of \muSplit in one glance:} Quantitative results of baselines \vs \muSplit variants.
    \figpart{a}~We plot the performance of the vanilla \UNet and the vanilla \HVAE baseline trained on increasingly larger patch sizes on our \PaviaATN Act \vs Tub data.
    The \UNet performance plateaus roughly at a patch size of about $256$.
    The performance of the vanilla \HVAE (not using LC) depends on how many hierarchy layers we use ($1$ to $4$, different colored plots), but then plateaus as well, or requiring a tremendous amount of GPU memory (black plot, also see Table~\ref{tab:microscopy_performance}).
    \figpart{b}~The left plot displays the data as shown in the \HVAE plot in (a), but now as a function of  hierarchy levels in the used architecture.
    Each curve is now representing a given patch size.
    X-axis ticks express how many hierarchy levels the \HVAE has, and how many of those make use of LC (number in brackets).
    The rightmost two plots show results obtained with \muSplit using an \HVAE with a patch size of only $64$.
    Each plot shows results obtained with one of our LC variations being used. 
    Not only do networks using LC outperform all baselines, they do so already when using the smallest patch size (64), thereby requiring only a moderate amount of GPU memory (see Table~\ref{tab:microscopy_performance}).
    }
    \label{fig:plots}
\end{figure*}
}

\newcommand\figShowImageResultsSmall{
\begin{figure*}[t]
\centering
\begin{overpic}[width=0.95\textwidth,tics=5]{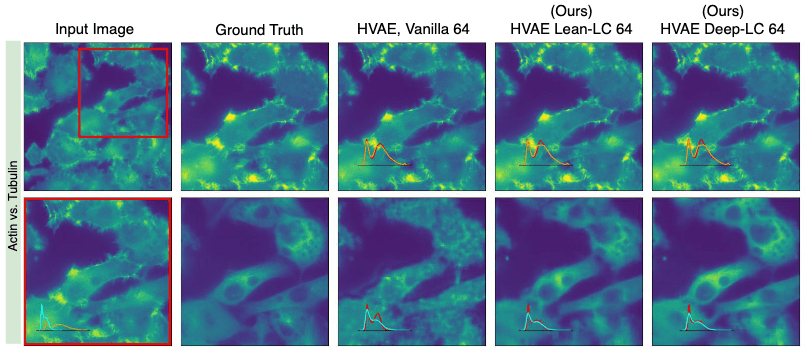}
 \put (42,37) {\tiny \color{white}{PSNR:28.3 (27.2)}}
 \put (61.5,37) {\tiny \color{white}{PSNR:30.4 (29.8)}}
 \put (81,37) {\tiny \color{white}{PSNR:30.1 (30.7)}}
 \put (42,  18) {\tiny \color{white}{PSNR:22.3 (23.1)}}
 \put (61.5,18) {\tiny \color{white}{PSNR:24.4 (25.5)}}
 \put (81,  18) {\tiny \color{white}{PSNR:24.0 (26.5)}}
\end{overpic}
\caption{Qualitative results on the Act vs.\ Tub task from our \PaviaATN dataset.
We compare ground truth to results obtained with the vanilla \HVAE baseline trained with a patch size of $64$ to results obtained with two variations of \muSplit (\HVAEs using lean and deep LC, both also using a patch size of $64$).
The overlaid histograms shows either the intensity distribution of the two channels (column 1) or the intensity distribution of the ground truth and the prediction (red).
The given PSNR are for the individual prediction (full input image) and for the entire dataset (in brackets).
}
\label{fig:image_results_small}
\end{figure*}
}

\newcommand\figReceptiveField{
\begin{figure*}[t]
\centering

\begin{overpic}[width=0.95\textwidth, tics=5]
            {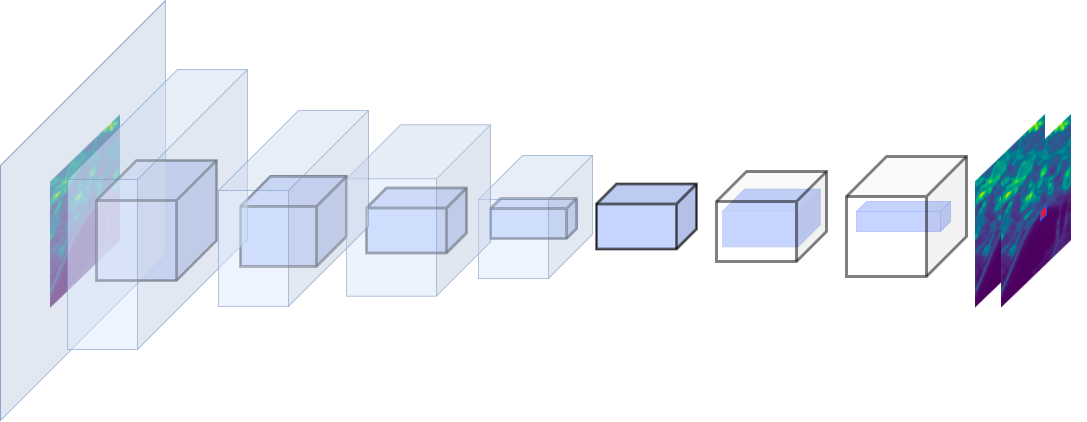}
\end{overpic}

\caption{
Cartoon of a generic hierarchical network with an encoder-decoder architecture illustrating the relationship between the input patch size, the effective receptive field, and the theoretical receptive field. 
The \textit{input patch}, shown at the very left in the center of the light blue area, is processed and downsampled multiple times (encoder) before being upsampled multiple times (decoder) to allow the output, shown on the very right, to have the same pixel dimensions as the input patch.
Cuboids shown by solid black lines represent the tensors the network computes during its execution.
Solid blue cuboids show the \textit{effective receptive field}, \ie the areas within each tensor that can influence the center-most pixels in the two output layers (depicted by red rectangles).
All but the last two tensors are fully `visible' to those pixels, since the \textit{theoretical receptive field}, \ie the maximum area that would influence those pixels if the respective tensor would be sufficiently large, grows beyond their bounds (shown as light-blue solid cuboids).
Note that working with larger input patches will fill a larger portion of the theoretical receptive field.
If theoretical and effective receptive fields diverge, as shown in this cartoon, padded predictions on input patches larger then the training patch size will cause the network to operate out-of-distribution (OOD) and therefore lead to degraded prediction quality (see main text and Supp.~Section~S.2.1).
}
\label{fig:receptive_field}
\end{figure*}
}

\newcommand\tabSyntheticPerformance{
\begin{table}[]
    \centering
    \begin{tabular}{c|c|c|c|c|c}
    Image & Model & \multicolumn{2}{c|}{$N_{join}=0$} & \multicolumn{2}{c}{$N_{join}=25$} \\
    Size  & & PSNR & SSIM & PSNR & SSIM \\
        \hline
        \multirow{3}{*}{128} & Vanilla & 28.3 & 0.90& 25.5 & 0.85 \\
         & \leanLC & \textbf{37.3} & 0.97 & 35.1 & 0.96 \\
         & \regularLC & 37.0 & \textbf{0.98} & \textbf{39.2} & \textbf{0.98} \\
        \hline
        \multirow{3}{*}{256} & Vanilla & 19.4 & 0.75 & 15.8 & 0.43 \\
        & \leanLC & 34.1 & 0.97 & 32.2 & 0.97\\
        & \regularLC & \textbf{41.5} & \textbf{0.99} & \textbf{41.6} & \textbf{0.98}  \\
    \end{tabular}
    \vspace{1mm}
    \caption{Quantitative results on the \SinosoidalCritters dataset.
    We compare results obtained with vanilla \HVAEs that do not use LC, and \HVAEs employing either \leanLC or \regularLC (\ie \muSplit results, see main text for details).
    All experiments are performed using a patch size of $64$.
    Bold numbers denote the best result for any given task (columns), showing that our results consistently outperform the vanilla baselines.
    }
    \label{tab:critters}
\end{table}
}

\newcommand\tabMicroscopyPerformance{
\begin{table*}[t]
    \centering
    \resizebox{.95\textwidth}{!}{%
    \begin{tabular}{l l r |r|r|r|r|r|r|r|r|r|r}
        \multicolumn{3}{l|}{} & & \multicolumn{6}{|c}{\PaviaATN} & \multicolumn{2}{|c}{\HagenEtAl}\\
        \cline{5-12}
        \multicolumn{3}{c|}{Model + Patch Size} & GPU & \multicolumn{2}{c|}{Act vs Nuc} & \multicolumn{2}{c|}{Tub vs Nuc} &  \multicolumn{2}{c}{Act vs Tub} & \multicolumn{2}{|c}{Act vs Mit}  \\
         \multicolumn{3}{l|}{} & (GiB) & PSNR & SSIM & PSNR & SSIM & PSNR & SSIM & PSNR & SSIM \\
         \hline
         \multicolumn{2}{l}{Double-DIP~\cite{Gandelsman2019-nn}} & & - & 22.8 & 0.30 & 21.2 & 0.20 & 20.9 & 0.30 &25.3 & 0.56 \\
         \multicolumn{2}{l}{BraveNet~\cite{Hilbert2020-ds}}            & 64 & 2.8 & 31.7 & 0.73 & 30.3 & 0.61 & 25.9 & 0.62 & 33.0 & 0.92\\
         \multicolumn{2}{l}{Context-Aware U-Net~\cite{Leng2018-ro}} & 64 & 4.7 & 31.5   & 0.74 & 29.0 & 0.61 & 25.1 & 0.61 & 31.1 & 0.91\\
         \hline
         \multicolumn{2}{l}{U-Net} & 256      &  9.4 & 33.2 & 0.79 & 31.4 & 0.71 & 28.1 & 0.69 & 34.2 & \textbf{0.95} \\
         \multicolumn{2}{l}{U-Net} & 512      & 28.7 & 33.3 & 0.79 & 31.1 & 0.72 & 27.9 & 0.69 & 34.1 & 0.94\\
         \hline
         U-Net & Regular-LC                 & 64      & 12.5   & 33.5 & 0.79 & 32.0 & 0.71 & 27.6 & 0.68 & 32.7 & 0.93\\
         \hline
         
         \multirow{5}{*}{\HAE}& Vanilla & 64     & 2.3  & 31.7 & 0.74 & 29.5 & 0.64 & 25.4 & 0.63 & 31.9 & 0.92 \\
         & Lean-LC & 64                          & 3.9  & 33.6 & 0.78 & 31.9 & 0.70 & 27.7 & 0.67 & 32.9 & 0.94  \\
         & Regular-LC & 64                               & 6.0   & 33.5 & 0.79 & 31.6 & 0.71 & 27.9 & 0.68  & 33.4 & 0.94  \\
         & Deep-LC & 64                          & 6.9 & 33.7 & 0.80 & 31.8 & 0.72 & 28.3 & 0.69 & 32.8 & 0.94 \\
         &Vanilla-XL  & 512                         & 31.2 & 33.2 & 0.79 & 30.2 & 0.68 & 27.6 & 0.67 & 34.2 & \textbf{0.95} \\
         \hline
         
         \multirow{5}{*}{\HVAE}& Vanilla & 64    & 2.8  & 31.8 & 0.75 & 29.6 & 0.64 & 25.2 & 0.61 & 31.9 & 0.93   \\          
         &                      Lean-LC & 64     & 4.4  & 33.8 & 0.79 & 31.9 & 0.71 & 27.7 & 0.68 & 32.7 & 0.94   \\ 
         &                      Regular-LC & 64          & 11.1 & \textbf{33.9} & 0.80 & 32.1 & 0.72 & 27.8 & 0.68 & 34.1 & \textbf{0.95}  \\ 
         &                      Deep-LC & 64     & 12.8 & \textbf{33.9} & \textbf{0.81} & 32.5 & \textbf{0.73} & \textbf{28.6} & \textbf{0.70}    & \textbf{34.3} & \textbf{0.95}\\
         &Vanilla-XL & 512 & ${(^*)}$& 33.4& 0.78 & \textbf{32.9} & 0.69 & 27.6 & 0.67 & \textbf{34.3} & \textbf{0.95}\\
        \hspace{1em}
    \end{tabular}
    }
    \caption{Quantitative results on fluorescent image decomposition tasks derived from the \PaviaATN and \HagenEtAl datasets.
    All results are reported in terms of peak signal-to-noise ratio (PSNR) and structural similarity index measure (SSIM).
    For each model we also report the used training patch size and GPU memory usage during training.
    The baselines we use are
    Double-DIP~\cite{Gandelsman2019-nn}, BraveNet~\cite{Hilbert2020-ds}, Context-Aware U-Net~\cite{Leng2018-ro},
    as well as vanilla \HAEs and \HVAEs using four hierarchy levels.
    Additionally, we show results for \UNets~\cite{Ronneberger2015-pk}, \HAEs, and \HVAEs trained on much larger patch sizes ($256$ and $512$).
    The results of \muSplit are also obtained with the same \HAE and \HVAE architectures trained on patches of size $64\times 64$, but with all hierarchy levels also employing either \leanLC, \regularLC, or \deepLC (see main text for details).
    Bold numbers denote the best result for any given task (column).
    In all but one case (\PaviaATN, Tubulin \vs Nuclei), our results outperform all baselines despite having a comparatively lean memory footprint. 
    Note that the Vanilla-XL \HVAE with patch size of $512$ and batch size of $32$ did not fit in 32 GiB of GPU memory and so we lowered the batch size such that the model did fit in memory. 
    }
    \label{tab:microscopy_performance}
\end{table*}

}

\newcommand\tabPatchsizeOptimalLC{
\begin{table}
\centering
\begin{tabular}{c|c|c|c}
     BU Blocks & vanilla 64 & rLC 64 & rLC 128 \\
     \hline
     1 &  24.3  & 24.7 & \textbf{24.8} \\
     2 &  25.1  & \textbf{25.9} & \textbf{25.9}\\
     3 & 25.2 & \textbf{27.0} & \textbf{27.0}\\ 
     4 & 25.4 & 27.8 & \textbf{27.9}
\end{tabular}
\vspace{1mm}
  \caption{Performance of \HVAE + \regularLC trained with patch size of $64$ (col 3) and $128$ (col 4) on Act vs Tub data.
  The larger patch size shows diminishing returns, indicating that LC is providing enough image context, showcasing the value of our approach.
  }
  \label{tab:patchsize_optimality_lc}
\end{table}
}

\newcommand\figPlotsActNuc{
\begin{figure*}[tb]
    \centering
    \begin{overpic}[width=.38\textwidth]{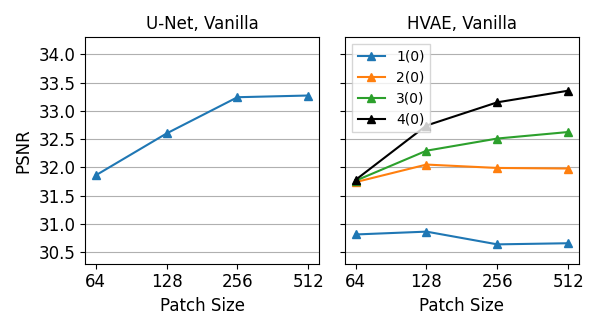}
        \put (0, 50) {\figpart{a}}
    \end{overpic}
    \hspace{5mm}
    \begin{overpic}[width=.58\textwidth]{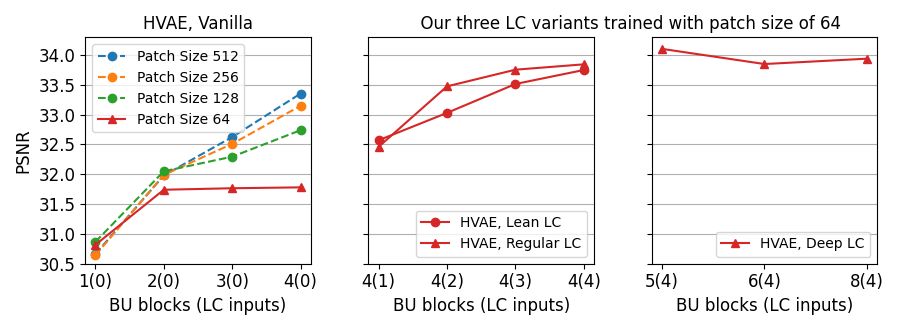}
        \put (0, 32) {\figpart{b}}
        \put(35.1,15.7){\tikz \draw[-to,dashed,red](0,0)--(0.5,0.4);}
        \put(66.8,29.2){\tikz \draw[-to,dashed,red](0,0)--(0.5,0.12);}

    \end{overpic}
    \caption{\textmathbf{Benefits of \muSplit in one glance:} Quantitative results of baselines \vs \muSplit variants on our \PaviaATN Act \vs Nuc task. In Figure~6 of the paper, we do the identical analysis on our \PaviaATN Act \vs Tub task. 
    \figpart{a}~We plot the performance of the vanilla \UNet and the vanilla \HVAE baseline trained on increasingly larger patch sizes.
    The \UNet performance plateaus roughly at a patch size of about $256$.
    The performance of the vanilla \HVAE (not using LC) depends on how many hierarchy layers we use ($1$ to $4$, different colored plots), but then plateaus as well, or requiring a tremendous amount of GPU memory (black plot, also see Table 1.
    \figpart{b}~The left plot displays the data as shown in the \HVAE plot in (a), but now as a function of  hierarchy levels in the used architecture.
    Each curve is now representing a given patch size.
    X-axis ticks express how many hierarchy levels the \HVAE has, and how many of those make use of LC (number in brackets).
    The rightmost two plots show results obtained with \muSplit using an \HVAE with a patch size of only $64$.
    Each plot shows results results obtained with one of our LC variations being used. 
    Not only do networks using LC outperform all baselines, they do so already when using the smallest patch size (64), thereby requiring only a moderate amount of GPU memory (see Table~1). Note that here, \deepLC has lesser advantage as was the case with Act vs Tub data.
    }
    \label{fig:plotsActNuc}
\end{figure*}
}

\newcommand\figInnerOuterPadding{
\begin{figure*}[t]
\centering

\begin{overpic}[width=.90\textwidth, tics=5] 
            {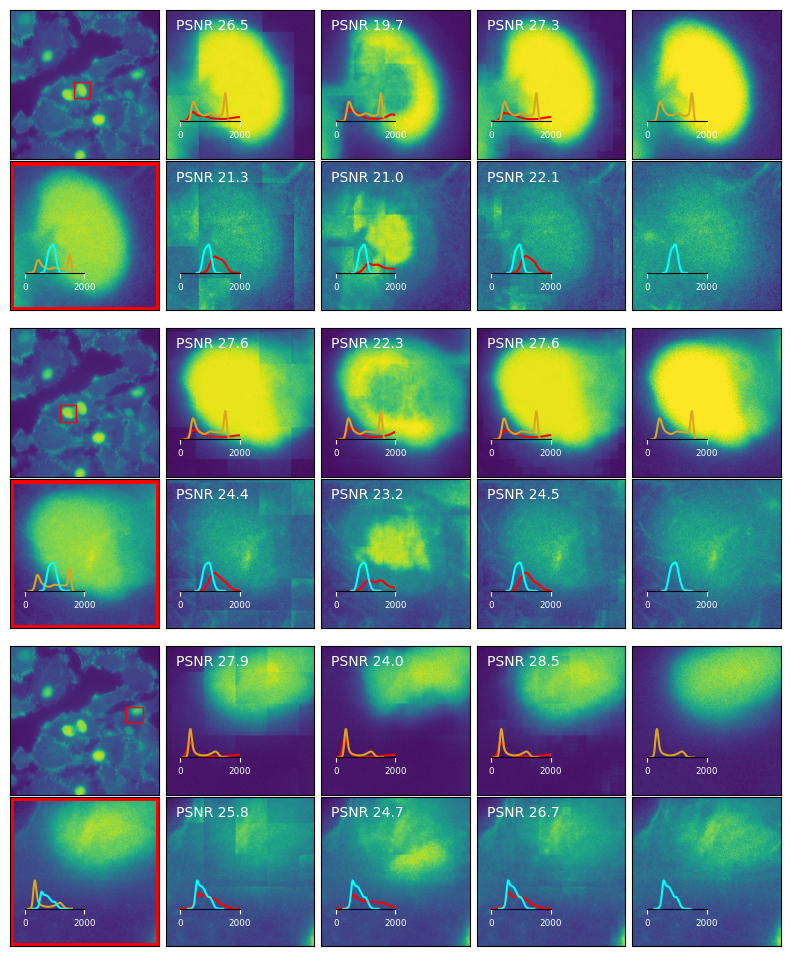}
        
        \put (5,100) {Input Image}
        \put (21.5,100) {No-padding}
        \put (36,100) {Outer-padding}
        \put (52.6,100) {Inner-padding}
        \put (72,100) {GT}
        \put (82.5,90) {Ch1}
        \put (82.5,75) {Ch2}

        \put (82.5,57) {Ch1}
        \put (82.5,42) {Ch2}

        \put (82.5,24) {Ch1}
        \put (82.5,9) {Ch2}

\end{overpic}

\caption{
Comparing \textit{No Padding}, \textit{Inner Padding} and \textit{Outer Padding} on the Actin vs Nucleus task. Here, we disentangle the region present inside the red square (column one) using Vanilla \HVAE model. One can see square shaped artefacts arising during tiling of predictions in 'No Padding' case, a case where no padding is used during the tiling operation. But more importantly, use of 'Outer Padding' leads to quite inferior splitting results where pixel intensity which was supposed to be present in the nuclues region 'leaks' into the actin channel (unexpected bright region in actin channel's prediction). This naturally degrades the splitting results of both channels. Inner tiling, on the other hand yields most consistent splitting results with respect to the ground truth (last column).
}
\label{fig:inner_outer_padding}
\end{figure*}
}

\newcommand\figRandomCropsAvsT{
\begin{figure*}[t]
\centering

\begin{overpic}[width=.99\textwidth, tics=5]
            {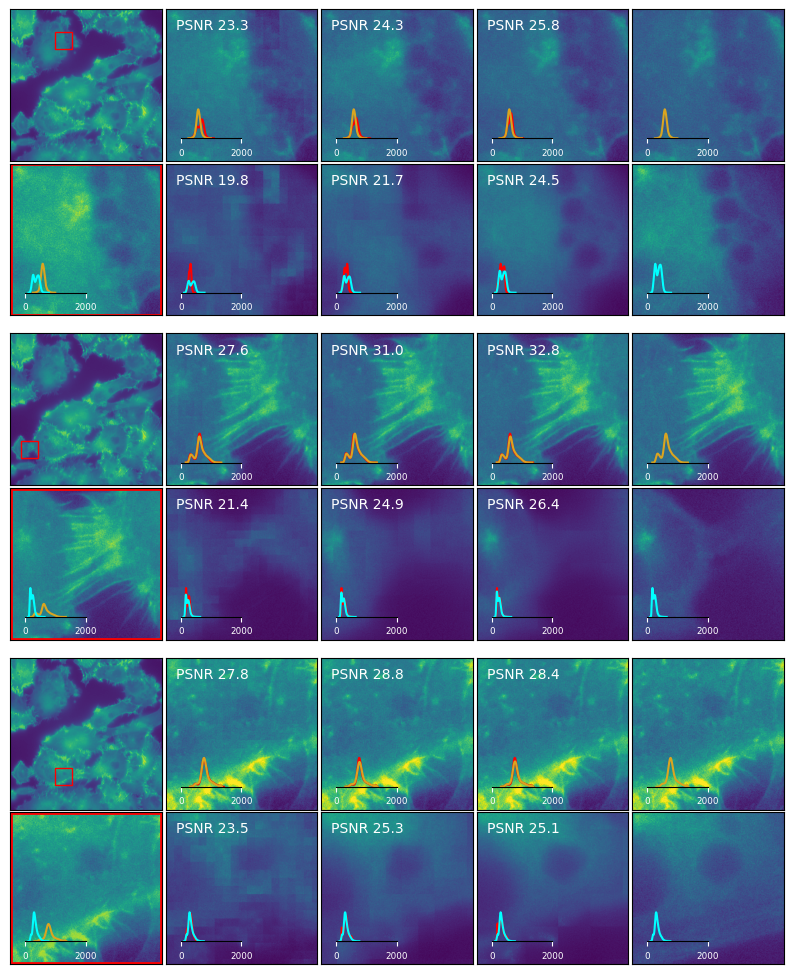}
        \put (3,100) {Input Image}
        \put (22,100) {Vanilla}
        \put (37,100) {Lean-LC}
        \put (52,100) {Deep-LC}
        \put (71,100) {GT}
        \put (81,90) {Ch1}
        \put (81,75) {Ch2}

        \put (81,57) {Ch1}
        \put (81,42) {Ch2}

        \put (81,24) {Ch1}
        \put (81,9) {Ch2}
    
\end{overpic}

\caption{
Qualitative evaluation of Vanilla \HVAE and our LC variants (also integrated to \HVAE architecture) on Actin vs Tubulin task. Here, we show results on three random crops of size $300\times300$. We disentangle the region inside red square, which is shown in column one. Last column has the ground truth for both channels.
}
\label{fig:randcrops_at}
\end{figure*}
}

\newcommand\figRandomCropsTvsN{
\begin{figure*}[t]
\centering

\begin{overpic}[width=.99\textwidth, tics=5]
            {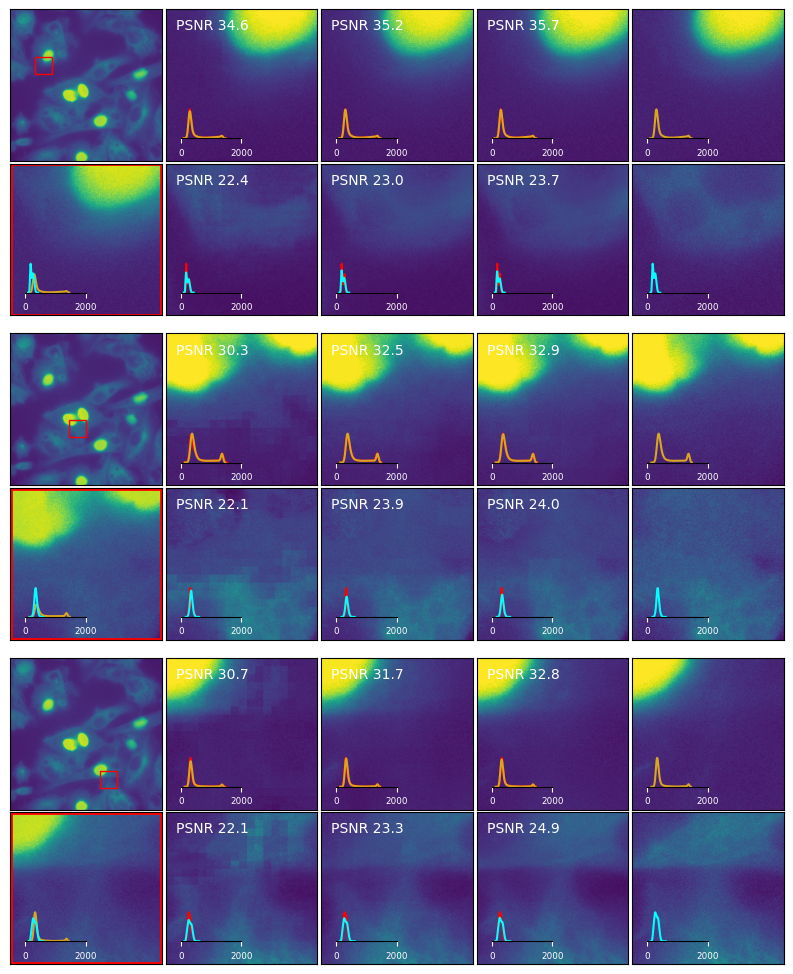}
        \put (3,100) {Input Image}
        \put (22,100) {Vanilla}
        \put (37,100) {Lean-LC}
        \put (52,100) {Deep-LC}
        \put (71,100) {GT}
        \put (81,90) {Ch1}
        \put (81,75) {Ch2}

        \put (81,57) {Ch1}
        \put (81,42) {Ch2}

        \put (81,24) {Ch1}
        \put (81,9) {Ch2}

\end{overpic}

\caption{
Qualitative evaluation of Vanilla \HVAE and our LC variants (also integrated to \HVAE architecture) on Tubulin vs Nucleus task. Here, we show results on three random crops of size $300\times300$. We disentangle the region inside red square, which is shown in column one. Last column has the ground truth for both channels.
}
\label{fig:randcrops_tn}
\end{figure*}
}
\newcommand\figRandomCropsAvsN{
\begin{figure*}[t]
\centering

\begin{overpic}[width=.99\textwidth, tics=5]
            {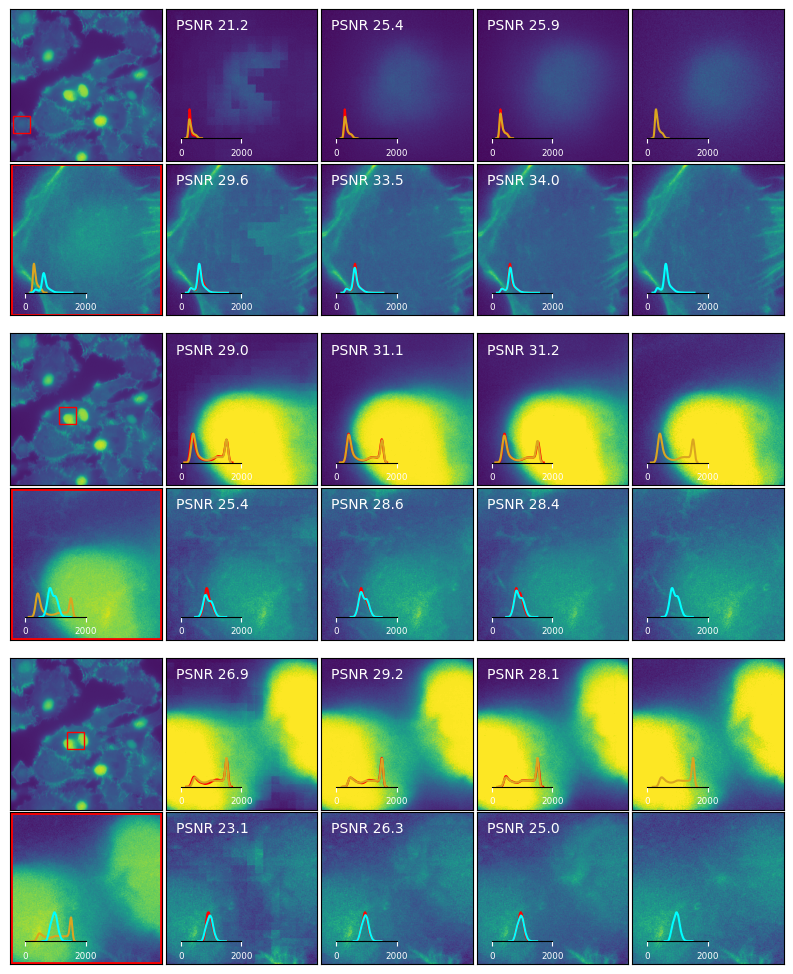}
        \put (3,100) {Input Image}
        \put (22,100) {Vanilla}
        \put (37,100) {Lean-LC}
        \put (52,100) {Deep-LC}
        \put (71,100) {GT}
        \put (81,90) {Ch1}
        \put (81,75) {Ch2}

        \put (81,57) {Ch1}
        \put (81,42) {Ch2}

        \put (81,24) {Ch1}
        \put (81,9) {Ch2}

\end{overpic}

\caption{
Qualitative evaluation of Vanilla \HVAE and our LC variants (also integrated to \HVAE architecture) on Actin vs Nucleus task. Here, we show results on three random crops of size $300\times300$. We disentangle the region inside red square, which is shown in column one. Last column has the ground truth for both channels.
}
\label{fig:randcrops_an}
\end{figure*}
}

\newcommand\figRandomImagesCrittersZero{
\begin{figure*}[b]
\centering

\begin{overpic}[width=.90\textwidth,  tics=5]{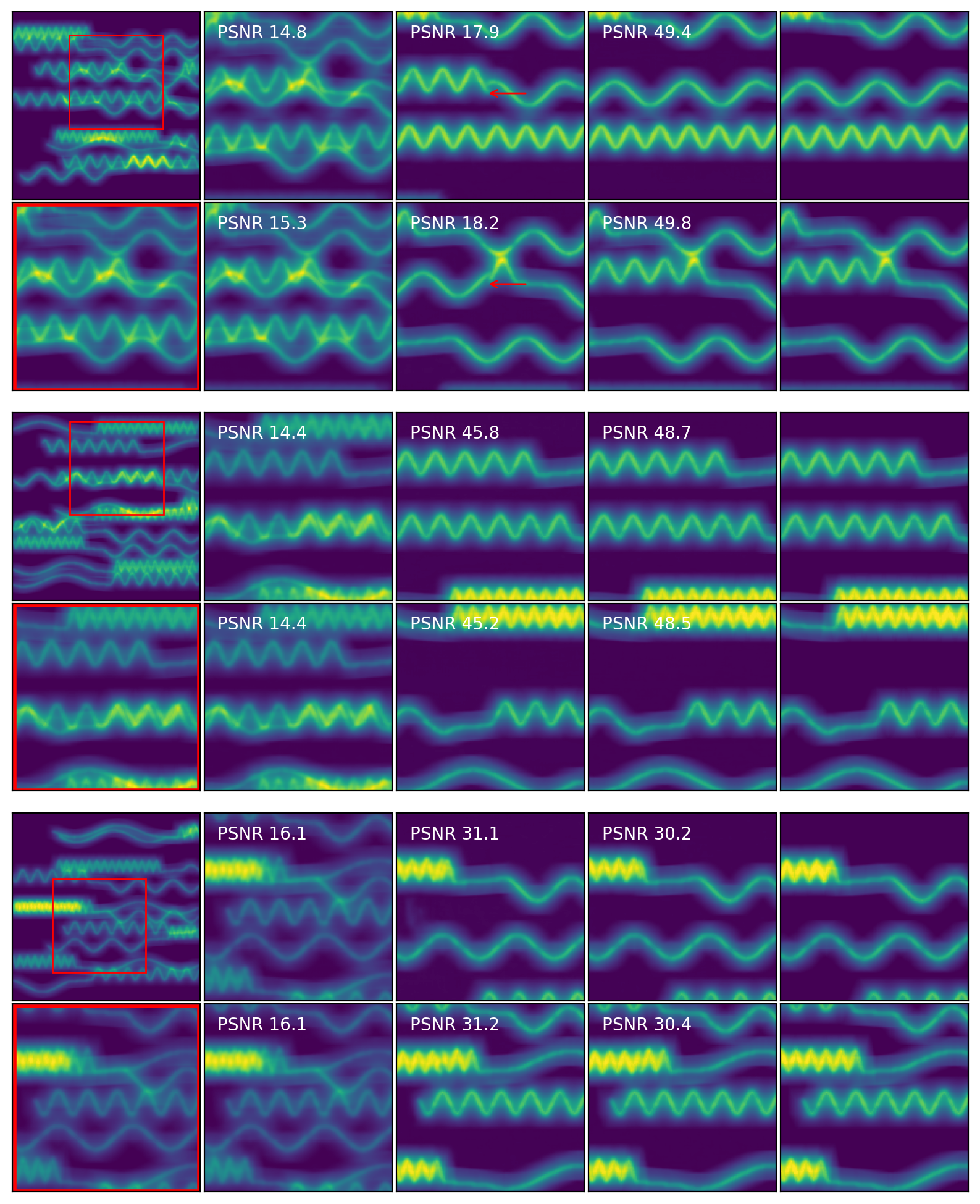}
        \put (3,100) {Input Image}
        \put (22,100) {Vanilla}
        \put (37,100) {Lean-LC}
        \put (52,100) {Deep-LC}
        \put (71,100) {GT}
        \put (81,90) {Ch1}
        \put (81,75) {Ch2}

        \put (81,57) {Ch1}
        \put (81,42) {Ch2}

        \put (81,24) {Ch1}
        \put (81,9) {Ch2}
    
\end{overpic}

\caption{
Qualitative evaluation of Vanilla \HVAE and our LC variants (also integrated to \HVAE architecture) on SinosoidalCritters dataset. Here, we show results on three random crops of size $200\times200$. We disentangle the region inside red square, which is shown in column one. Last column has the ground truth for both channels. Red arrows highlight few interesting areas where we observe our \deepLC performs better than others.
}
\label{fig:randcrops_critters0}
\end{figure*}
}

\newcommand\figLeanArchitecture{
\begin{figure*}[b]
\centering
    \begin{overpic}[width=0.99\textwidth, tics=2] 
            {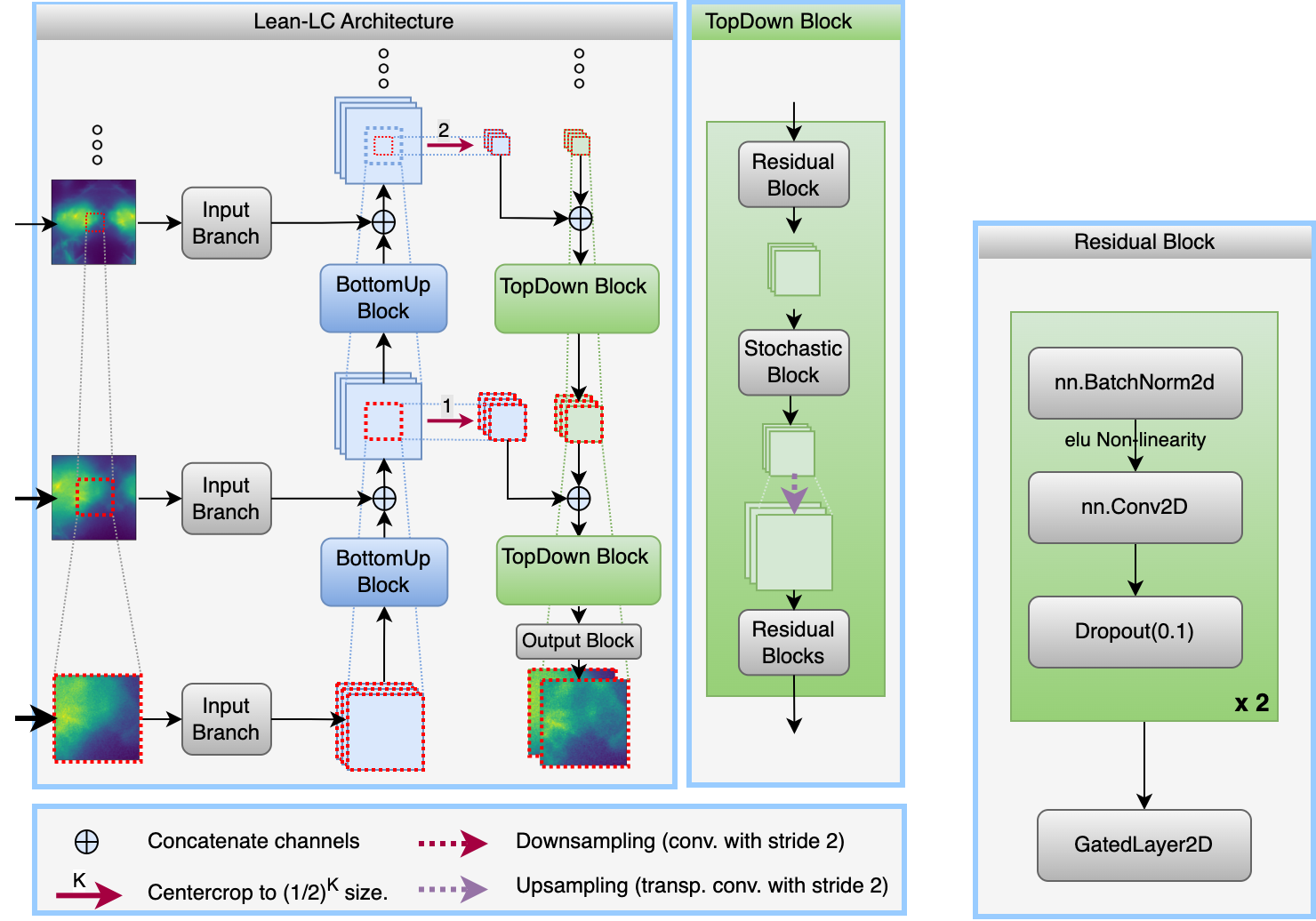}
        \put (64.6, 67.6) {\cite{Prakash2020-wr}}
        \put (42, 45.5) {\cite{Prakash2020-wr}}
        \put (42, 24.9) {\cite{Prakash2020-wr}}
        \put (92.3, 51) {\cite{Prakash2020-wr}}
        \put (-0.5, 68) {\figpart{a}}
        \put (71, 51) {\figpart{b}}

    \end{overpic}

\caption{ Lean-LC and Residual block architecture.
\figpart{a} The network architecture of Lean-LC variant is shown here. 
The Bottom-Up block remains unchanged from the architecture of LC and it is Top-Down block which has changed. 
If we look at $k^{th}$ Bottom-Up block (from bottom), then as before, the output from the $k^{th}$ Bottom-Up block is passed to the next Bottom-Up block and also to the Top-Down block to the same level. 
However, before feeding to the Top-Down block, the output is center-cropped to $(1/2^{k})^{\text{th}}$ size.
The Top-Down block of LC-Lean is identical to the Top-Down block of~/cite{Prakash2020-wr}. 
Input to the block passes through Residual blocks and then through the stochastic block. 
The output of the Stochastic block is upsampled to twice its size through Transposed convolution with stride of 2.
~\figpart{b}The schema for the Residual Block. 
This is directly taken from~\cite{Prakash2020-wr}.
}
\label{fig:overall_architecture_lean}
\end{figure*}
}

\newcommand\figDoubleDip{
\begin{figure}
    \centering
    \begin{overpic}[width=0.45\textwidth,tics=5]{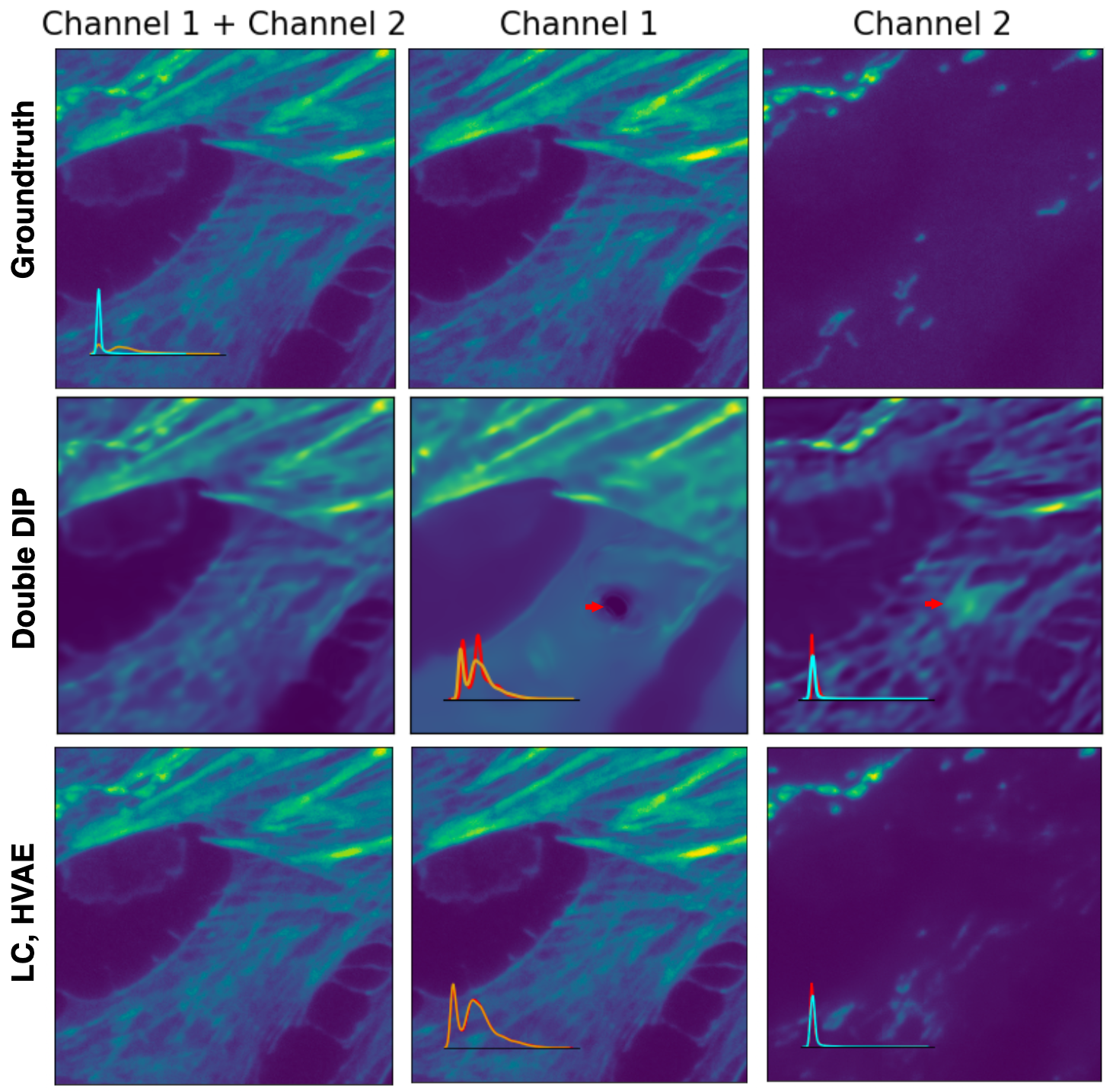}
     \put (18,59) {\tiny \color{white}{PSNR:32.8 (28.5)}}
     \put (50,59) {\tiny \color{white}{PSNR:23.7 (25.7)}}
     \put (81,59) {\tiny \color{white}{PSNR:23.5 (24.9)}}
     \put (18,27) {\tiny \color{white}{PSNR:49.7 (36.1)}}
     \put (50,27) {\tiny \color{white}{PSNR:32.8 (30.0)}}
     \put (81,27) {\tiny \color{white}{PSNR:30.5 (31.8)}}
    \end{overpic}
    
    \caption{Qualitative image decomposition results using the \DoubleDIP baseline (row 2) on an $256\times256$ image crop from \HagenEtAl dataset.
    The overlaid histograms shows either the intensity distribution of the two channels (column 1) or the intensity distribution of the ground truth and the prediction (red). 
    \regularLC, on the other hand, performs well.
    Note that \DoubleDIP is solving a much harder task since it is an unsupervised method trained on a single input images.
    }
    \label{fig:doubledip_imgs}
\end{figure}
}

\newcommand\figRandomCropsAvsM{
\begin{figure*}[t]
\centering

\begin{overpic}[width=.99\textwidth, tics=5]
            {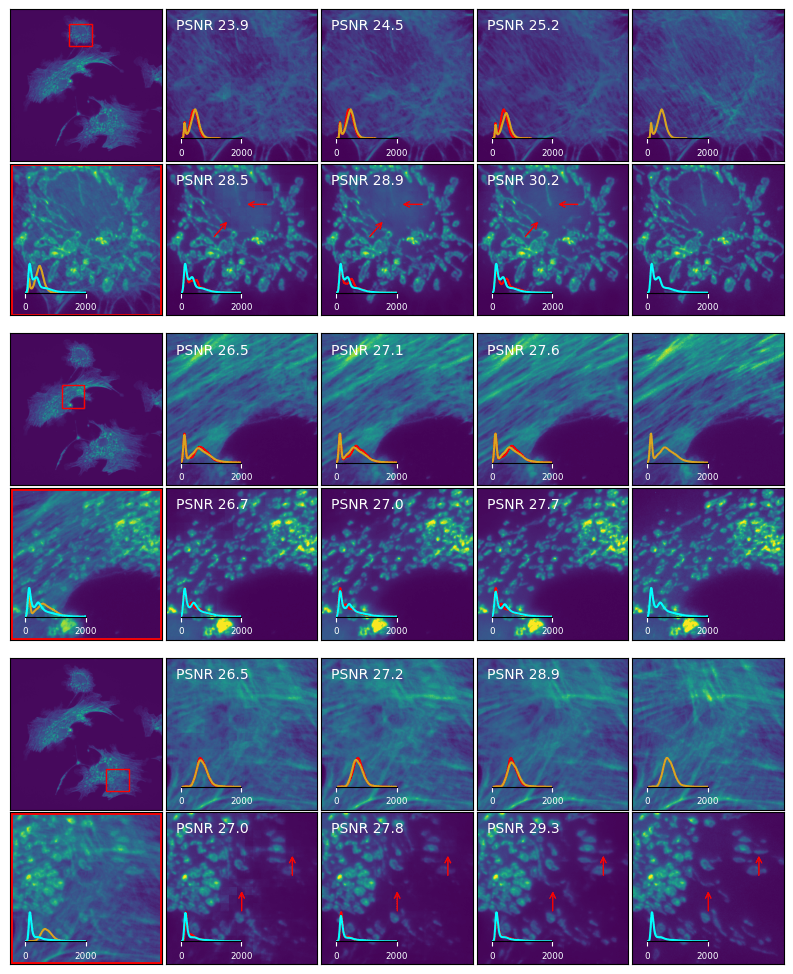}
        \put (3,100) {Input Image}
        \put (22,100) {Vanilla}
        \put (37,100) {Lean-LC}
        \put (52,100) {Deep-LC}
        \put (71,100) {GT}
        \put (81,90) {Ch1}
        \put (81,75) {Ch2}

        \put (81,57) {Ch1}
        \put (81,42) {Ch2}

        \put (81,24) {Ch1}
        \put (81,9) {Ch2}

\end{overpic}

\caption{
Qualitative evaluation of Vanilla \HVAE and our LC variants (also integrated to \HVAE architecture) on Actin vs Mitochondria task. Here, we show results on three random crops of size $300\times300$.
Input to all models is the region inside red square, as seen in column one. Last column has the ground truth for both channels. Red arrows highlight few interesting areas where we observe our \deepLC performs better than others.
}
\label{fig:randcrops_am}
\end{figure*}
}

\begin{document}

\title{$\bm{\mu}$Split: image decomposition for fluorescence microscopy}

\author{
	Ashesh$^{1}$, Alexander Krull$^{2}$, Moises Di Sante$^{3}$, Francesco Silvio Pasqualini$^{3}$, Florian Jug$^{1,}$\thanks{Corresponding Author, (florian.jug@fht.org).} \\
	\normalsize{$^1$Human Technopole, Italy}, \normalsize{$^2$University of Birmingham, UK},
	\normalsize{$^3$University of Pavia, Italy}\\
        \tt\small {ashesh.ashesh@fht.org, a.f.f.krull@bham.ac.uk, moises.disante@unipv.it}\\
        \tt\small{francesco.pasqualini@unipv.it, florian.jug@fht.org}
}
\maketitle
\ificcvfinal\thispagestyle{empty}\fi

\begin{abstract}
    We present \muSplit, a dedicated approach for trained image decomposition in the context of fluorescence microscopy images.
    We find that best results using regular deep architectures are achieved when large image patches are used during training, making memory consumption the limiting factor to further improving performance.
    We therefore introduce lateral contextualization (LC), a novel meta-architecture that enables the memory efficient incorporation of large image-context, which we observe is a key ingredient to solving the image decomposition task at hand.
    We integrate LC with U-Nets, Hierarchical AEs, and Hierarchical VAEs, for which we formulate a modified ELBO loss.
    Additionally, LC enables training deeper hierarchical models than otherwise possible and, interestingly, helps to reduce tiling artefacts that are inherently impossible to avoid when using tiled VAE predictions.
    We apply \muSplit to five decomposition tasks, one on a synthetic dataset, four others derived from real microscopy data.
    Our method consistently achieves best results (average improvements to the best baseline of 2.25 dB PSNR), while simultaneously requiring considerably less GPU memory. Our code and datasets can be found at \href{https://github.com/juglab/uSplit}{https://github.com/juglab/uSplit}. 
    
\end{abstract}

\section{Introduction}
\label{sec:intro}

Fluorescence microscopy~\cite{Ghiran2011-np} is routinely used to look at living cells and biological tissues at cellular and sub-cellular resolution~\cite{Ouyang2022-fj}. 
Components of the imaged cells can be highlighted using fluorescent labels, allowing biologists to investigate individual structures of interest.
Given the complexity of biological processes, it is typically necessary to look at multiple structures simultaneously, typically via a temporal multiplexing scheme~\cite{Ghiran2011-np} that separates them into different image channels.
\figTeaser

Imaging more than 3 or 4 structures in this way is difficult for technical reasons, limiting the rate of scientific progress in the life sciences.
One way to circumvent this limitation would be to label two cellular components with the same fluorophore, \ie image them in the same image channel.
Hence, a computational method to split apart (decompose) superimposed biological structures acquired in a single image channel, \ie without temporal multiplexing, would have tremendous impact (see Figure~\ref{fig:teaser}).

Historically, image decomposition has found applications on natural images~\cite{Gandelsman2019-nn,Dekel2017-os,Bahat2016-qp,Berman2016-nz}. 
Our approach for image decomposition, called \muSplit, rests on the idea of learning structural priors for the two unmixed target image channels, and then using these to guide the decomposition of the superimposed (added) pixel intensities.
Such content-aware priors have previously been used for tasks such as 
image restoration~\cite{Weigert2018-pi, Buchholz2019-kk, Weigert2017-hc}, 
denoising~\cite{Krull2018-cj, Batson2019-ae, Krull2020-my, Goncharova2020-jz, Prakash2020-wr, Prakash2021-dz}, and 
segmentation~\cite{Buchholz2020-vd, Schmidt2018-is, Weigert2019-ih}. 

In many of these cases, the achievable performance heavily depends on the portion of the image a network can see before having to make a prediction. 
As we show in this work, the need for large spatial context, \ie receptive field and patch size, is particularly pronounced for image decomposition.  
Biological structures in microscopy images can easily extend over distances of several hundred pixels. 
Accordingly, we observe that results improve with larger training patch sizes and deeper architectures (see Figure~\ref{fig:plots}(a)). 
Naturally, this leads to models having a huge GPU memory footprint, which limits their applicability to selected compute-savvy life-science labs. 

The importance of context has previously been utilized in the field of image segmentation~\cite{Leng2018-ro,Hilbert2020-ds}.
Leng~\etal~\cite{Leng2018-ro} devised a method to efficiently use the available context of the input image for a segmentation task. 
However, they did not use additional inputs for having access to a larger context than what is already present in the given input patch.
Hilbert~\etal~\cite{Hilbert2020-ds} worked with 3D images and used an additional lower resolution image to improve overall segmentation performance. 

Also for \muSplit we observe that additional image context is important.
In contrast to the previously mentioned architectures, we introduce Lateral Contextualization (LC), a novel meta-architecture that feeds additional image context at multiple processing steps.
We introduce three variants, \leanLC, \regularLC, and \deepLC, differing from each other in terms of GPU memory requirements and achievable prediction quality.
As we elaborate below, \deepLC additionally offers the possibility to instantiate a more powerful \HierarchicalVAE with more hierarchy layers than otherwise possible, and show that this leads to improved performance on the image splitting task at hand.
e
Since \muSplit needs to be applicable to large microscopy images, tiled predictions are required.
In tiled predictions, input image is divided into overlapping patches on which predictions are performed individually.
Those predictions are then appropriately center-cropped into non-overlapping tiles which can then be appended to form the final prediction. 
Overlapping patches have to be used to ensure that sufficient image context is available to address border artifacts to occur in the non-overlapping central region.

In Section~\ref{sec:padding}, we argue that for deep networks operating on relatively small patches, overlapping regions should not be created by making tiles larger (Outer Padding) which is arguably the most common way, but that it is better  to instead center-crop regions smaller than the original patch size (Inner Padding).

Since \HierarchicalVAEs (\HVAEs)~\cite{Sonderby2016-lq} have recently gained popularity, \eg for microscopy image denoising and restoration~\cite{Prakash2020-wr,Prakash2021-dz}, we made these powerful architectures also available to the image decomposition task by modifying the default VAE ELBO loss, incorporating the fact that the fed input is different from the decoded output. 

\section{Problem Statement}
\label{sec:problem_statement}
A dataset $D_{mix}=(x^1, x^2,..,x^N) $ of $N$ images is created by superimposing sampled pairs of image channels $(D_1,D_2)$, such that
\begin{equation}
    x^i = (d_1^i + d_2^i)/2, \forall i \in [1,N],
    \label{eq:d_mix}
\end{equation}
with $D_1=(d_1^1, d_1^2,... d_1^N)$ and $D_2=(d_2^1,d_2^2...,d_2^N)$.

Given a newly sampled $x = (d_1 + d_2)/2$, the task is to decompose $x$ into estimates of $d_1$ and $d_2$. 

\figArchitecture
\section{Our Approach}
\label{sec:methods}

\miniparagraph{A Sound ELBO for $\bm{\mu}$Split}
\label{sec:ELBO}
We train our VAE to describe the joint distribution for both channel images $d_1$ and $d_2$.
We modify the VAE's ELBO objective to incorporate the fact that input and output are not the same (as they are for autoencoders). 
When training the VAE, our objective is to find 
\[
\argmax_{\decparams} \:
\sum_{i=1}^N
\log P(d^i_1,d^i_2;\decparams),
\]
based on our training examples $(d^i_1,d^i_2)$. 
Here, $\decparams$ are the decoder parameters of our VAE, which define the distribution.
Next, we expand $\log P(d_1,d_2;\decparams)$ as
\[
\log \int P(d_1,d_2,z;\decparams) dz
\]
\[
=  \log \int q(z|x;\encparams)*
\frac{P(d_1,d_2,z;\decparams)}{q(z|x;\encparams)} dz
\]
\begin{equation}
    >=\int q(z|x;\encparams) * \log \frac{P(d_1,d_2,z;\decparams)}{q(z|x;\encparams)}dz,
    \label{eq:elbo}
\end{equation}
where $q(z|x;\encparams)$ is our encoder network with parameters $\encparams$.
It can be shown that the evidence lower bound in Eq.~\ref{eq:elbo} is equal to
\[
E_{q(z|x;\encparams)}[\log P(d_1,d_2|z;\decparams)] -KL(q(z|x;\encparams),P(z)).
\]
By making the assumption of conditional independence of $d_1$ and $d_2$ given $z$, we can simplify the expression to 
\begin{equation}
    \begin{split}
        \:E_{q(z|x;\encparams)} [
        \log P(d_1|z;\decparams) + \log P(d_2|z;\decparams)
        ] \\
        -KL(q(z|x;\encparams),P(z)).
    \end{split}
    \label{eq:DVAE_exp_final}
\end{equation}
Expression~\ref{eq:DVAE_exp_final} is what we end up maximizing during training. 
Note that this analysis can be seamlessly extended to the case where one has a hierarchy of latent vectors~\cite{Sonderby2016-lq} instead of just one. 

For modelling $q(z|x;\encparams)$, we use the identical setup of the bottom-up branch used in~\cite{Prakash2020-wr} with the input being $x$, the superimposed input. 
For modeling $P(d_1|z;\decparams)$ and $P(d_2|z;\decparams)$, we again use the top-down branch design used in~\cite{Prakash2020-wr} but make the top-down branch output two channels for mean and two more for the pixelwise $\log(var)$, one each for $d_1$ and $d_2$. 
So, the output of our model is a 4 channel tensor with identical spatial dimensions as the input. 
Note that to encorporate LC, we modify both $q(z|x;\encparams)$ and $P(d_2|z;\decparams)$ which we describe next.

\miniparagraph{Lateral Contextualisation (LC)}
\label{sec:lc}
We introduce LC, allowing \muSplit to see large portions of the input image at increasingly downscaled pixel resolutions. 
LC only requires small full resolution patches, rendering the network considerably more memory efficient. 

Many popular architectures, such as \UNets~\cite{Ronneberger2015-pk} or \HVAEs~\cite{Prakash2020-wr,Child2020-ya,Vahdat2020-hk} are composed of a hierarchy of levels that operate on increasingly downsampled and therefore also increasingly smaller layers.
The basic idea of LC is to pad each downsampled layer by additional image context, \ie additional input from an available larger input image, such that each layer at each hierarchy level maintains the same spatial dimensions.
(In Figure~\ref{fig:overall_architecture} (a), the red dashed squares in the stack of inputs (leftmost column) indicate the location of the original patch ($x_p$) within the downscaled and laterally contextualized inputs at higher hierarchy levels ($x_{(p,i)}$).)

\miniheadline{Creating downsampled LC inputs} 
Let $x_p=x_{[c,h]}$ denote a patch of size $h\times h$ from $x \in D_{mix}$ centered around pixel location $c$. 
To decompose the patch $x_p$, we additionally use a sequence of successively downscaled and cropped versions of $x$, 
$X_p^\text{lowres} = (x_{(p,1)},x_{(p,2)},\dots, x_{(p,n_{\text{LC}}}))$, 
where $x_{(p,k)}$ is $x_{[c,2^k\cdot h]}$, downsampled to the same pixel resolution of $h\times h$, and $n_{\text{LC}}$ denotes the total number of used LC inputs.

\miniparagraph{Implementation of \regularLC}
Overall architecture is shown in Figure~\ref{fig:overall_architecture}(a). Primary input patch $x_p$ is fed to the first input branch (IB).
The output of this IB is fed to the first bottom up (BU) block, which downsamples the input via strided convolutions, whose output is then 
passed to some residual blocks (see Supp. figure~\textit{S.1}), and finally zero padded to regain the same spatial dimension as the input it received. 
The output of the first BU block is concatenated with the output of the second IB, which has received the first lower resolution input containing additional lateral context, $x_{(p,1)}$. 
Zero-padding followed by concatenation ensures pixelwise alignment between IB's output and BU's output. 
We use $1\times1$-convolutions to merge these concatenated channels and feed the resulting layer into the next BU block.
This procedure gets repeated for every hierarchy level in the given \HVAE.
\figReceptiveField

Once the topmost hierarchy level is reached, the last layer is fed into the topmost top down (TD) block.
A TD block consist of some residual layers, followed by a stochastic block as they are used in \HVAEs.
The output of the stochastic block is center-cropped to half size and upsampled via transpose convolutions before again being fed through some residual layers ((see Supp. figure~\textit{S.1})).
Cropping and upsampling ensures that the output of the TD block matches the next lower hierarchy level.
The output of the TD block is, similar to before, first concatenated with the output of the bottom up computations and then fed through $1\times 1$-convolutions.
Once we reach the bottom hierarchy level, the output of the last TD block is fed through an output block (OB) composed of some additional convolutional layers, giving us the final predictions of $d_1$ and $d_2$. 

We've integrated LC into \HVAE, \HAE and the classic \UNet architecture. 
Note that the difference between \HVAEs and \HierarchicalAutoencs~(\HAEs) is that the stochastic block is replaced by the identity.
We use the term \textit{Vanilla} to denote the underlying architecture on which we then enable LC. 

\figPaddingFigures

\miniparagraph{\deepLC: deeper performs better}
We observe empirically that having deeper hierarchies is beneficial (see Figure~\ref{fig:plots}(a)). 
Since in \UNets, \HAEs, and \HVAEs, each consecutive hierarchical level halves the input tensor in all spatial dimensions, a natural limit to the maximum hierarchy level is given by the fed patch size\footnote{Using a patch size of 64, for example, can at most give rise to 5 hierarchy levels ($2^{5+1} = 64$).}. 
By making use of additional lower resolution image context at each hierarchy level, we've designed \muSplit such that spatial dimensions of latent tensors stay constant across all hierarchy levels. 
This enables \deepLC (see Figure~\ref{fig:overall_architecture}(b)), our most potent architecture variant, to have additional hierarchy levels over what a vanilla \HVAE can have, typically showing best results in our experiments (see Figure~\ref{fig:plots}(b) and Figure~\ref{fig:image_results_small}).

More concretely, in our \deepLC network, we stack a default \HVAE (like the one used in~\cite{Prakash2020-wr}) on top of our \regularLC variant (Figure~\ref{fig:overall_architecture}(a)).
This means that starting from the highest hierarchy level using LC, any further hierarchy level is built like a regular \HVAE hierarchy stack. 

\miniparagraph{\leanLC: minimal memory footprint}
\leanLC, our most memory efficient LC variation, does not use the lateral context introduced in the bottom-up branch within the top-down branch (see Supp. Figure S.1 for its architecture).
More specifically, the bottom-up branch is identical to \regularLC, but the top-down branch reduces to the  default \HVAE implementation, very similar to how it was also used in~\cite{Prakash2020-wr}. 
This is enabled by centercropping the output of each BU block going into the TD block.

\figSynthetic
\miniparagraph{Tiled Predictions}
\label{sec:padding}
For virtually all tasks using fully convolutional architectures, trained networks are often used to predict results on inputs much larger then the patches they were trained on.
Whenever an input image is so large that the network in question cannot scale without running out-of-memory, predictions are typically performed on overlapping patches and later suitably cropped and appended. 
When applied to relatively shallow~\cite{ronneberger_u-net:_2015} and non-variational networks, results can be pixel-perfect, \ie not containing any tiling artifacts.
But we observe that there are two cases wherein tiling artefacts are not easily avoidable.

The first is caused by networks that have huge receptive fields (see Figure~\ref{fig:receptive_field}). 
When trained with a patch size much smaller than the theoretical receptive field size, large parts of the theoretical receptive field will be empty (\ie zero). 
See also Supp.~Section~S.2.1 for a more detailed description.

When such trained networks are later used for tiled predictions, a problem arises whenever the input patches, on which predictions are made, are larger than the patch size used during training (which typically is the case because patch sizes is chosen such that GPU memory is best utilized, and input patches need to overlap sufficiently to avoid border artifacts). 
These patches will fill a larger portion of the theoretical receptive field than training patches did, resulting in out-of-distribution (OOD) predictions and worsened performance (see Figure~\ref{fig:padding_figures} (b) for quantitative assessment). 

The second case for tiling artifacts arises when variational networks like \HVAEs are used. 
These architectures sample from the variational latent space of encoded tiles, with samples for neighboring tiles not necessarily decoding into consistent image contents along the borders of predicted tiles.

The solution we propose is twofold: 
$(i)$~Instead of tiled prediction on large patches (Outer Padding), which is arguably the most often used tiling scheme, we propose to use Inner Padding instead, an approach that uses patches of the same size as the ones used during training, thereby solving the OOD issue introduced above. 
More specifically, in both tiling schemes, the input image is divided into overlapping patches. 
The predictions on these patches are then centercropped and these crops are put right next to each other in order to create a prediction for the entire input image. 
To enlarge the overlap between neighboring patches, Outer Padding enlarges the patch size. 
Inner Padding does not alter the size of patches, but instead only uses a smaller central area of their respective predictions. 
See Figure~\ref{fig:padding_figures}~(a) for a visual depiction of Inner and Outer Padding.
In our experiments (see Section~\ref{sec:experiments}), we have used Inner Padding of $24$ pixels, determined via grid-search.
$(ii)$~Overlap amount with Inner Padding are constrained to be small. Small overlap would usually cause artifacts due to insufficient image context at tile boundaries. 
However, due to our LC approach, \muSplit is fed a very large and consistent image context at both sides of all patch boundaries, allowing us to operate with minimal artifacts even with small overlaps\footnote{Note that artifacts arising from independently sampling the latent space in \HVAEs remains an unsolved problem.}. In supplement, we empirically show the lower need of overlap for our LC variants.

\miniparagraph{Training Details}
\label{sec:training}
For every dataset, we use 80\%, 10\% and 10\% of the data as train-validation-test split.
All models are trained using 16-bit precision on a Tesla V100 GPU. 
Unless otherwise mentioned, all models are trained with batch size of $32$ and input patch size of $64$. 
For all \HVAEs, we lower-bound $\sigma$s of $P(d_1,d_2,\theta)$ to $\exp({-5})$. 
This avoids numerical problems arising from these $\sigma$s going to zero, as reported in \cite{Rezende2018-yn}.
Next, we re-parameterize the normal distributions for the BU branch using $\sigma_{ExpLin}$ reformulation introduced in~\cite{Dehaene2021-ix}. 
We additionally upper-bound the input to $\sigma_{ExpLin}$ to $20$.
For training \muSplit with \deepLC, we follow the suggestions in~\cite{Child2020-ya,Radford2019-wg}, and divide the output of each BU block by $\sqrt{2\dot i}$, with $i$ being the index of the hierarchy level the BU block is part of.

\figPlots
\figShowImageResultsSmall

\section{Datasets}
\label{sec:data}
\miniparagraph{SinosoidalCritters}

We created this synthetic dataset explicitly to demonstrate the importance of context for the splitting task and the usefulness of using LC within \muSplit. 
Images in this dataset can only correctly be decomposed when sufficient lateral image context is available during prediction time.

We first choose 4 different frequencies and combine them into 4 unique pairs.
Two pairs are dedicated for image channel 1 (blue box), the other two for image channel 2 (green box). 
We call these pairs \textit{critters}. The assignment of these critters to channels is done such that each frequency is assigned exactly once to each channel.
We connect the two sinosoids of each critter with a low frequency curve of controllable length ( later denoted by $N_{\text{join}}$ in Table~\ref{tab:critters}). Note that it is the specific combination of sinosoid frequencies present in the curve which decides whether it belongs to Channel 1 or 2 since the individual sinosoids themselves occur in both channels in equal amount.
Next, we assemble channel images by placing a predefined number of randomly chosen curves at random positions in the respective image channel.
The final input image is created as the sum of the two channels. See Figure~\ref{fig:critters}(a) for dataset construction.

\miniparagraph{\PaviaATN Microscopy Dataset} 
We've created \PaviaATN dataset. It has been imaged in the Synthetic Physiology Laboratory at University of Pavia, and is composed of 62 4-channel fluorescence microscopy images of size $2720\times 2720$. Notably, this dataset has higher pixel resolution than most publicly available fluoroscence microscopy datasets~\cite{Hagen2021-xh,ounkomol_label-free_2018,zhang_poisson-gaussian_2019}.
The three channels we use label Actin, Tubulin and Nuclei, respectively, yielding three decomposition tasks we refer to as Actin \vs Tubulin, Actin \vs Nuclues, and Tubulin \vs Nucleus.
Note that the dataset has two channels labelling Nuclei from which we picked one. 
See supplement for more details.

\miniparagraph{\HagenEtAl Actin-Mitochondria Dataset}
From many sub-datasets provided by Hagen and colleagues~\cite{Hagen2021-xh}, we picked the one with Mitochondria and Actin channels, the one with the highest pixel resolution ($2048\times2048$).

\tabMicroscopyPerformance
\tabSyntheticPerformance

\tabPatchsizeOptimalLC
\section{Experiments and Results}
\label{sec:experiments}
\label{sec:results}
\miniparagraph{Incrementally Introducing LC}
\label{introducting_lc}
In left panel of Figure~\ref{fig:plots}(b), we show that for Vanilla \HVAE, as hierarchy levels increase (BU blocks), so does the performance, provided we've large enough patch size. For patch size of 64, increasing hierarchy levels does not bring any benefit after a point. 

In central panel of Figure~\ref{fig:plots}(b), keeping the patch size and hierarchy levels fixed to 64 and 4 respectively, we introduce LC to an increasing number of hierarchy levels (denoted by the number in the brackets along the x-axis). 
This gives us a cumulative gain of around 2dB PSNR. Furthermore, with \deepLC (right panel), we increase the hierarchy level even further which gives us further improvements. 
Two things are worth noting here for the patch size of 64: 
$(i)$~There is not much benefit in increasing hierarchy levels for Vanilla \HVAE. Using LC, on the other hand, leads to additional improvements, and
$(ii)$~Vanilla \HVAE, cannot employ as many hierarchy levels as we can do using \deepLC, and the results gain substantially from those extra levels.  
The Vanilla-XL model denotes Vanilla model trained with a patch size of 512. 
The \deepLC results outperform the Vanilla-XL \HVAE, see Figure~\ref{fig:plots}(a), while also having a much smaller GPU memory footprint (see Table~\ref{tab:microscopy_performance}).

\miniparagraph{Experiments on Microscopy Data}
\label{sec:results_real_data}
We present results on $3$ decomposition tasks on the \PaviaATN dataset and $1$ decomposition task on the \HagenEtAl dataset. 
Table~\ref{tab:microscopy_performance} summarizes our findings.
As baselines, we've adapted the works of~\cite{Leng2018-ro,Hilbert2020-ds} and find that \muSplit outperforms them. 
It is worth noting that architecture used in~\cite{Hilbert2020-ds}, unlike ours, did not generalize to using a hierarchy of lower resolution inputs and worked with just one additional low resolution input.
It also, unlike us, did not respect pixel alignments while concatenating the latent space tensors of the two resolution levels. 
We have also applied the unsupervised Double-DIP~\cite{Gandelsman2019-nn} baseline to random sampled $6$ crops of size $256\times 256$ for each test-set image of the \PaviaATN and \HagenEtAl datasets (see Table~\ref{tab:microscopy_performance} and supplementary figure).

Over all four tasks, the best performing LC variant with \HVAE architecture outperforms the best LC variant with \HAE architecture by 0.5 PSNR on average.
Using the \HVAE architecture, \deepLC outperforms \leanLC on average by $0.8$ PSNR. 
For the \HAE architecture this difference is $0.1$ PSNR.
Qualitative results are shown in Figure~\ref{fig:image_results_small} and in the supplement.
%

\miniparagraph{Outer \vs Inner Padding and Runtime Performance} 
In Figure~\ref{fig:padding_figures}(c), we show the percentage change in PSNR with different amounts of padding and see that the vanilla \HAE and \HVAE setup performances degrade (left plot) when Outer Padding is used with large padding amounts. 
But with Inner Padding (right plot), we see improvement saturation with increase in padding amount.  
In Figure~\ref{fig:padding_figures}(b), one can observe an artefact appearing solely due to Outer Padding 
(artifact does not exist in 'No padding'). 
These results support our claim about OOD issue as described in Section~\ref{sec:methods}.

Note that Inner Padding requires a larger number of individual predictions, indicated by the smaller grid size seen in Figure~\ref{fig:padding_figures}(a) (denoted by red dashed rectangle).
Specifically, using an Inner Padding of $24$ pixels with a patch size of $64$ will use $16\times 16$ center-crop per patch. 
Hence, we will need to predict $16$ ($(64/16=4)^2$) times more patches to cover the entire input image.

Interestingly, we found padding giving minor benefits for \deepLC quantitatively and so \deepLC results in Table~\ref{tab:microscopy_performance} were computed without padding thereby leading to a better runtime for \deepLC. 
However, we still find few tiling artefacts with \deepLC and in those cases Inner Padding helps. Other two LC variants benefit both quantitatively and qualitatively from Inner Padding.

\miniparagraph{Effects of Larger Training Patch Sizes}
In Figure~\ref{fig:plots}(a) we show that increasing the training patch size improves the performance of a \UNet and vanilla \HVAEs across different hierarchy levels.
While the \UNet baseline performance saturates, \HVAEs' improvement with increasing hierarchy levels does not, but quickly reach a hard limit in terms of GPU memory requirement (see Table~\ref{tab:microscopy_performance}).

\miniparagraph{Performance of LC with larger patch sizes}
Using \muSplit, microscopy labs having limited GPU compute will still get similar performance to labs with ample resources, labs capable of using networks employing larger patch sizes. So far, all our LC variants have been trained with a patch size of $64$. 
A natural question to ask is whether there is still some benefit in using larger patch sizes when also using LC. 
While the answer to this question depends upon multiple factors like how much long range interactions are present in the data, the receptive field size of the network etc, we did an ablation to empirically investigate this in Table~\ref{tab:patchsize_optimality_lc}. 
One can observe that for \HVAE + \leanLC, across different hierarchy levels (BU Block count), using a patch size of $128$ only provides a minor performance improvement over a patch size of $64$. 
This implies that for a pixel's prediction, only a small amount of neighbourhood context needs to be given at native pixel resolution and most of the context can be given via lower-resolution lateral image context. 


\miniparagraph{Experiments on Synthetic Data}
\label{sec:results_synthetic_data}
In Table~\ref{tab:critters} we show the results obtained on the \SinosoidalCritters dataset.
We used two input image sizes, $128\times 128$ and $256\times 256$, and two values for $N_{join}$, namely $0$ and $25$ pixels. 
On average, \muSplit outperforms the vanilla \HVAE by $18$ PSNR. 
Also note that the larger input size, constituting a harder problem to solve, is resulting in a drop of performance for the vanilla \HVAE.
Using \muSplit, instead, the performance increases.
To recognise which critter is depicted and assign it to a channel, the network has to see both wave forms. 
The vanilla \HVAE is able to do splitting on $128\times 128$, but it has artefacts (red circle in Figure~\ref{fig:critters}(b)). 
For the $256\times 256$ pixel images, it completely fails because it is unable distinguish between the critters since it cannot simultaneously process a sufficiently large part of the image.
In contrast, by using LC we are able to successfully split both images.

\miniparagraph{\UNet Hyperparameter Tuning}
\label{sec:baseline_tuning}
We tuned depth and patch size of a classic \UNet to achieve optimal performance for the tasks at hand (see supplement for details). 



\section{Discussion}
\label{sec:discussion}
In this work, on our dataset we show that \muSplit performs better when deeper architectures, \ie \HAEs and \HVAEs, are employed and enabled to process additional image context via the memory efficient lateral contextualization (LC) schemes we propose.

The deeper such networks become, the larger will the receptive field (RF) sizes grow, in our case routinely exceeding sizes of $512\times 512$ pixels.
An immediate consequence of this is that we cannot easily employ common tiling schemes (\ie Outer Padding) without running into out-of-distribution (OOD) issues (see Section~\ref{sec:methods}).
Hence, we propose to use Inner Padding to circumvent this problem.
Additionally, we observe that \deepLC does even perform quite well without padded tiled predictions (no additional overlap between patches).
The reason for this is that the patch context typically given by overlapping regions is now substituted by context being fed via \deepLC.
Still, best performance is typically obtained using \deepLC and Inner Padding during tiled predictions.

It is important to point out that for any variational models, such as \HVAEs, tiled predictions suffer from the additional problem that neighboring tiles will likely not be consistent due to the sampling step performed independently per tile.
While Inner Padding still is the better strategy to employ (for the same argument as for any other model with huge receptive fields), sampling inconsistencies cannot be fully avoided.
The strength of these artifacts will depend on the data uncertainty (\ie the ambiguity in the fed inputs \wrt the trained model).

In summary, we have proposed a powerful new method to efficiently use image context.
We have then applied this method to an impactful new image decomposition task on fluorescence microscopy data.
We believe that the presented ideas will prove to also be useful in the context of other computer vision problems.
We will explore the applicability of LC to other problem domains in future work. 
Additionally, we will make \muSplit more amenable to noisy fluorescence data and to disentanglement tasks where more than two image channels are superimposed.

\section*{Acknowledgements}
\label{sec:ack}
This work was supported by 
the European Commission through the Horizon Europe program (IMAGINE project, grant agreement 101094250-IMAGINE and AI4LIFE project, grant agreement 101057970-AI4LIFE) as well as the compute infrastructure of the BMBF-funded de.NBI Cloud within the German Network for Bioinformatics Infrastructure (de.NBI) (031A532B, 031A533A, 031A533B, 031A534A, 031A535A, 031A537A, 031A537B, 031A537C, 031A537D, 031A538A).
Additionally, the authors also want to thank Damian Dalle Nogare of the Image Analysis Facility at Human Technopole for useful guidance and discussions and the IT and HPC teams at HT for the compute infrastructure they make available to us.


{\small
\bibliographystyle{ieee_fullname}
\bibliography{Paperpile_muSplit}
}

\newpage

\onecolumn
\begin{center}

  \textbf{\Large Supplementary Material\\$\bm{\mu}$Split: efficient image decomposition for microscopy data}\\[.2cm]
Ashesh$^{1}$,
Alexander Krull$^{2}$,
Moises Di Sante$^{3}$, \\
Francesco Silvio Pasqualini$^{3}$,
Florian~Jug$^{1}$ \\

\textsuperscript{1}Jug Group, Fondazione Human Technopole, Milano, Italy, 
\textsuperscript{2}University of Birmingham, United Kingdom, 
\textsuperscript{3} University of Pavia, Italy
\\
\end{center}
\setcounter{equation}{0}
\setcounter{figure}{0}
\setcounter{table}{0}
\setcounter{page}{1}
\renewcommand{\thepage}{S.\arabic{page}} 
\renewcommand{\thefigure}{S.\arabic{figure}}
\renewcommand{\thetable}{S.\arabic{table}}
\renewcommand*{\thesection}{S.\arabic{section}}
\setcounter{section}{0}
\figLeanArchitecture

\twocolumn
\section{The Architecture of Lean-LC}

In this section, we describe the \leanLC architecture, our most GPU memory efficient LC variation (see Supplementary Figure~\ref{fig:overall_architecture_lean}). 
In this architecture, lateral contextualization is only used along the bottom-up branch. 
The top-down branch therefore reduces to a regular \HVAE, in our case just as the one used in~\cite{Prakash2020-wr}. 
To make the laterally contextualized bottom-up branch funnel into the regular (vanilla) \HVAE top-down branch, the output of each BottomUp block feeding into the corresponding TopDown block is appropriately centercropped.
Hence, the latent tensors in the top-down branch are smaller, leading to the reduced memory footprint. 

\figPlotsActNuc

\section{Padding used in Tiling}
\subsection{Issue with Outer Padding}
\label{sec:issue_outer_padding}
Here we introduce two terminologies needed to explain the issue with Outer Padding. 
Assuming an infinitely large input or intermediate tensor, we define its \textit{theoretical receptive field} to be the subset of tensor entries which can influence a single output pixel (see Figure~3 of the paper).
Given a finite tensor size, governed by a fixed input patch size (\eg $64\times64$) we define the \textit{effective receptive field} analogously as the subset of tensor entries which can influence a single output pixel.
Note that the theoretical receptive field is either identical to the effective receptive field or larger (see Figure~3).

As we use a deep network and work with $64\times64$ sized input patches during training, the theoretical receptive field (with grows up to about $500\times$500) is much larger than the effective receptive field (which cannot grow beyond $64\times64$).
Given that the network has the capacity to see a large region but a much smaller patch is fed as input, a natural question to ask is: what does the network `see' beyond the input, \ie, its effective receptive field? 
The answer is that it sees  zeros due to zero padding present in \textit{same-convolution} operations of PyTorch.
Hence, the network is accustomed to see lot of zeros during training. 

At evaluation time, if we use Outer Padding, we increase the patch size, and therefore the effective receptive field. 
Now, suddenly, the network sees a lot fewer zeros and it is therefore not surprising that the quality of predictions degrades when so much more input is fed through the network.
In such cases, the network will effectively start operating out of distribution (OOD) with respect to the training data that was consistently fed at the same patch size of $64\times64$.

Importantly, when using Inner Padding, we do not change the patch size during tiled predictions and are therefore avoiding to operate the trained network OOD.  

\subsection{Qualitative Results for Inner Padding}
\figInnerOuterPadding
In Supplementary Figure~\ref{fig:inner_outer_padding}, we compare results obtained with Inner Padding, Outer Padding, and without padding.
For this we evaluated on random patches of size $400\times400$ from the Actin vs Nucleus dataset. 
One can easily spot square-shaped artefacts when no padding is used. 
With Inner Padding, they are generally much improved. 
Outer Padding also leads to a reduction of these artefacts, but generate other artefacts leading to degraded performance in the splitting task at hand.

On the full dataset, PSNR drops from 31.4~dB PSNR without padding to 30.2~dB with Outer Padding.
Using Inner Padding, on the other hand, improves the obtained results to a PSNR of 31.8~dB (+0.4~dB). 
\subsection{Deep-LC makes padding less important}
We compare results obtained with a vanilla \HVAE, our architecture using \leanLC, and our variant using \deepLC.
In Table~\ref{tab:padding_needs}, we report the PSNR we achieve on four datasets with all of the architectures, once using no padding during prediction, the other time using Inner Padding of 24 pixels.
On average, the performance when using Inner Padding improved by 0.33~dB PSNR for the vanilla \HVAE, 0.18~dB when using \leanLC, and 0.02~dB when using \deepLC. 
This supports our claim that \deepLC brings enough lateral context that padding becomes generally less important.

\begin{table}[]
    \centering
\begin{tabular}{c|c|c|c|c|c|c|}
        Dataset & \multicolumn{2}{c|}{Vanilla} & \multicolumn{2}{c|}{\leanLC} & \multicolumn{2}{c|}{\deepLC}\\
        & P0 & P24& P0 & P24& P0 & P24\\
        \hline
        Act vs Nuc  & 31.4 & 31.8 & 33.7 & 33.8 & 33.9 & 33.9\\
        Act vs Mito & 31.4 & 31.9 & 32.4 & 32.7 & 34.3 & 34.4\\
        Act vs Tub  & 25.0 & 25.2 & 27.5 & 27.7 & 28.6 & 28.6 \\
        Tub vs Nuc  & 29.4 & 29.6 & 31.8 & 31.9 & 32.5 & 32.5 \\
    \end{tabular}
    \vspace{1mm}
    \caption{LC can offset the importance of padding. 
    The table shows results by a vanilla \HVAE and our \muSplit with \leanLC and \deepLC.
    For each model, we show the achieved PSNR without padding (P0) and when Inner Padding of 24 pixels (P24) was used during prediction. 
    One can observe that the difference between the two columns for each architecture becomes increasingly smaller, suggesting that LC helps to avoid artefacts typically addressed with padding during tiled predictions.}
    \label{tab:padding_needs}
\end{table}

\section{The PaviaATN Data}
The PaviaATN dataset comprises static lambda-stacks from a human keratinocytes cell line (HaCaT) expressing GFP-tubulin, RFP-LifeAct, and a customized version of the cell cycle indicator FastFUCCI that uses various combinations of a yellow fluorescent protein (YPF, mTurquoise2) and a far-red fluorescent protein (iRFP, miRFP670) to indicate multiple phases of the cell cycle. 
While the details of how this cell line was genetically engineered will be published separately, here we used it to create a challenging dataset for \muSplit which we planning publish with this manuscript. 
In fact, when a cell is in the G1 phase, increasing intensities of YFP fluorescence are detected in the nucleus. 
As a cell moves from G1 to S phase (G1/S), both YFP and iRFP fluorescence are detected in the nucleus of the cell. 
Finally, the sole iRFP fluorescence is detected in the nucleus during the S-G2-M phase. 
At the onset of the G1 phase, the nucleus shows no visible fluorescence intensity. 
All Images are acquired through the 100x silicon oil objective of a Nikon Ti2 microscope (100x silicon oil objective) equipped with an Okolab environmental control chamber and a Crest V3 spinning disk confocal in widefield mode. 
Excitation light was provided by a Lumencor Celesta laser engine set up to provide 5\% laser power to the 446, 477, 546, and 637 nm lines. 
Emission light was collected through the following filters Semrock FF01-511/20, 595/31, 685/40. 
This configuration can spectrally separate the signals from actin (RFP) and S-G2-M cell cycle phases (iRFP). 
Instead, GFP and YFP exhibit a degree of overlap in excitation (446 and 477) and emissions (through the Semrock 595/31 filter), which we seek to resolve with \muSplit. 
Since all combinations of YFP and GFP variants have spectral overlap, we expect the \muSplit results to be very relevant for the field.

\section{Metrics}
We use PSNR and SSIM (structural similarity) to quantitatively measure the quality of predictions. 
When reporting PSNR, we use the commonly used shift invariant variation introduced in~\cite{Weigert2018-pi}. 
We compute both SSIM and PSNR metrics on normalized data.

\section{More Quantitative Results}
Figure~\ref{fig:plotsActNuc} shows the achievable results of a U-Net, a vanilla \HVAE, and our \muSplit variations for the PaviaATN Actin vs Nucleus data.
Plots are as the ones in the main figure. 
One can observe the outperformance of LC variants with respect to the Vanilla baseline. 
On this task we observe that the \deepLC architecture does not lead to additional improvements over the other LC variations.
This can be explained by the nucleus channel being relatively easy to separate from the actin channel, without requiring much lateral image context to perform the task well.

\section{More Qualitative Results}
\figRandomCropsAvsM
\figRandomCropsAvsT
\figRandomCropsTvsN
\figRandomCropsAvsN
\figRandomImagesCrittersZero


All qualitative results in Supplementary Figures~\ref{fig:randcrops_am},~\ref{fig:randcrops_at},~\ref{fig:randcrops_tn} and ~\ref{fig:randcrops_an} are showing predictions on randomly chosen patches of size $300\times300$. 

All qualitative results figures show randomly chosen patches in two rows, each one showing one of the superimposed image channels. 
The superimposed input region is shown in the first column and the last column shows ground truth. 
All other columns show predictions from various model configurations.  

Supplementary Figures~\ref{fig:randcrops_am},~\ref{fig:randcrops_at},~\ref{fig:randcrops_tn} and ~\ref{fig:randcrops_an} all show results for \HVAE variations, \ie comparing the vanilla architecture, with the ones utilizing \leanLC, and \deepLC.

In Supplementary Figure~\ref{fig:randcrops_critters0}, we show performance on random inputs of our SinosoidalCritters data.

\subsection{Comparison with Double-DIP}
In Supplementary Figure~\ref{fig:doubledip_imgs}, we qualitatively compare \muSplit's performance with Double-DIP's performance. We note that Double-DIP, being a completely unsupervised approach, naturally finds it difficult to know the 'correct' split, the split which exists in nature. It simply returns one of the many plausible splitting options. Its inferior performance argues for some form of supervision for our problem. 

\figDoubleDip

\section{Different Neural Network Submodules}
\paragraph{Residual Block} We've taken the residual block formulation from~\cite{Prakash2020-wr}. 
The schema for the residual block is shown in Supplementary Figure~\ref{fig:overall_architecture_lean}~\figpart{b}. 
The last layer in the residual block is the GatedLayer2D which doubles the number of channels through a convolutional layer, then use half the channels as gate for the other half.

\paragraph{Stochastic Block} The channels of the input of this block are divided into two equal groups. 
The first half is used as the mean for the Gaussian distribution of the latent space. 
The second half is used to get the variance of this distribution, implemented via the $\sigma_{ExpLin}$ reformulation introduced in~\cite{Dehaene2021-ix}.  

\section{U-Net Tuning}
We varied the depth of the used U-Net. 
For consistency with the other used architectures, we decided to still call it BottomUp (BU) blocks (\HAEs and \HVAEs grow upwards, not downwards.) 
Table~\ref{tab:unet_finetuning} shows the achievable performance with U-Nets of different depth (number of BU blocks).

\begin{table}[]
    \centering
    \begin{tabular}{c|c}
        BU Blocks & PSNR \\
        \hline
         1 & 29.8  \\
         2 & 31.3  \\
         4 & 33.2 \\
         5 & 33.2 \\
         6 & 33.0 \\
    \end{tabular}
    \vspace{1mm}
    \caption{The achievable performance using a U-Net using various numbers of bottom-up (BU) blocks. 
    For the results reported in the main text, 5 BU blocks have been used.}
    \label{tab:unet_finetuning}
\end{table}

Other relevant hyperparameter values used for U-Nets are $\text{patience} = 200$ for early stopping , $\text{patience} = 30$ for the learning rate scheduler (ReduceLROnPlateau).

\twocolumn
\end{document}